\documentclass[acmsmall]{acmart}

\usepackage{booktabs} 

\usepackage{times}
\usepackage{epsfig}
\usepackage{graphicx}
\usepackage{amsmath}
\usepackage{amssymb}

\usepackage{bm}
\usepackage{bbm}
\usepackage{amsfonts}
\usepackage{amssymb}
\usepackage{graphicx}
\usepackage{subfigure}
\usepackage{multirow}
\usepackage{booktabs}
\usepackage{diagbox}
\usepackage{makecell}
\usepackage{textcomp}
\usepackage{tgpagella}
\usepackage{lineno}
\usepackage{longtable}
\newcommand{\tabincell}[2]{\begin{tabular}{@{}#1@{}}#2\end{tabular}}

\usepackage[ruled]{algorithm2e} 

\SetAlFnt{\small}
\SetAlCapFnt{\small}
\SetAlCapNameFnt{\small}
\SetAlCapHSkip{0pt}
\IncMargin{-\parindent}

\AtBeginDocument{
  \providecommand\BibTeX{{
    \normalfont B\kern-0.5em{\scshape i\kern-0.25em b}\kern-0.8em\TeX}}}

\begin{document}
\title{Adaptive Exploration for Unsupervised Person Re-identification}

\author{Yuhang Ding}
\email{dyh.ustc.uts@gmail.com}
\affiliation{
  \institution{SUSTech-UTS Joint Centre of CIS, Southern University of Science and Technology, China}
}

\author{Hehe Fan}
\affiliation{
  \institution{Centre for Artificial Intelligence, University of Technology Sydney}
  \streetaddress{15 Broadway}
  \city{Ultimo, Sydney, NSW 2007}
  \country{Australia}}
\email{hehe.fan@student.uts.edu.au}

\author{Mingliang Xu}
\affiliation{
 \institution{School of Information Engineering, Zhengzhou University}
 \city{Zhengzhou, Henan}
 \country{China}}
\email{iexumingliang@zzu.edu.cn}

\author{Yi Yang}
\affiliation{
  \institution{Centre for Artificial Intelligence, University of Technology Sydney}
  \streetaddress{15 Broadway}
  \city{Ultimo, Sydney, NSW 2007}
  \country{Australia}}
\email{yee.i.yang@gmail.com}

\acmYear{2020}
\acmDOI{10.1145/3369393}

\acmJournal{TOMM}
\acmVolume{16}
\acmNumber{1}
\acmArticle{3}
\acmMonth{2}

\renewcommand{\shortauthors}{Y. Ding et al.}

\begin{abstract}
Due to domain bias, directly deploying a deep person re-identification (re-ID) model trained on one dataset often achieves considerably poor accuracy on another dataset.
In this paper, we propose an Adaptive Exploration (AE) method to address the domain-shift problem for re-ID in an unsupervised manner.
Specifically, in the target domain, the re-ID model is inducted to 1) maximize distances between all person images and 2) minimize distances between similar person images.
In the first case, by treating each person image as an individual class, a non-parametric classifier with a feature memory is exploited to encourage person images to move far away from each other.
In the second case, according to a similarity threshold, our method adaptively selects neighborhoods for each person image in the feature space.
By treating these similar person images as the same class, the non-parametric classifier forces them to stay closer.
However, a problem of the adaptive selection is that, when an image has too many neighborhoods, it is more likely to attract other images as its neighborhoods. 
As a result, a minority of images may select a large number of neighborhoods while a majority of images has only a few neighborhoods.
To address this issue, we additionally integrate a balance strategy into the adaptive selection.
We evaluate our methods with two protocols. 
The first one is called ``target-only re-ID'', in which only the unlabeled target data is used for training.
The second one is called ``domain adaptive re-ID'', in which both the source data and the target data are used during training.
Experimental results on large-scale re-ID datasets demonstrate the effectiveness of our method.
Our code has been released at \url{https://github.com/dyh127/Adaptive-Exploration-for-Unsupervised-Person-Re-Identification}.
\end{abstract}

\ccsdesc[500]{Computing methodologies}
\ccsdesc[500]{Computing methodologies~Visual content-based indexing and retrieval}
\ccsdesc[500]{Computing methodologies~Image representations}

\keywords{person re-identification, unsupervised learning, domain adaptation, deep learning}

\maketitle

\section{Introduction}\label{sec:introduction}

Person re-identification (re-ID) has become more and more popular because it plays an important role in security. 
Given an image of a person-of-interest from one camera, the goal of person re-ID is to find the person from other cameras. 
In recent years, benefiting from deep convolutional neural networks (CNNs) and large-scale datasets, person re-ID has made great progress.
However, due to different cloth styles, camera viewpoints, scenes, \textit{etc.}, deep re-ID models often suffer from the problem of domain shift.
Specifically, when we directly apply a pre-trained deep re-ID model to an unseen dataset, it often achieves considerably poor accuracy. 

To address the domain-shift problem, one can collect a large amount of labeled data from the target domain for supervised training.
However, it can be costly and time-consuming to collect a large-scale dataset.
We can also collect a small amount of labeled data and a large amount of unlabeled data from the target domain for semi-supervised training \cite{DBLP:journals/tip/WuLDYBY19, DBLP:conf/iccv/LiuWL17}.
However, when the number of target domains increases, this kind of methods are still impractical.
Therefore, a promising solution is to use unlabeled data in target domains to train deep re-ID models, which is referred to as unsupervised person re-ID \cite{lin2019aBottom, DBLP:conf/cvpr/Deng0YK0J18, DBLP:conf/cvpr/WeiZ0018, DBLP:journals/tomccap/FanZYY18, DBLP:conf/cvpr/WangZGL18, DBLP:conf/bmvc/LinLLK18, DBLP:journals/tip/ZhongZZLY19, DBLP:journals/corr/abs-1904-01990}.
This paper concerns unsupervised person re-ID.

There are two lines of unsupervised re-ID methods. 
The first line is to directly fine-tune a deep CNN model (usually pre-trained on ImagNet \cite{DBLP:conf/cvpr/DengDSLL009}) on the unlabeled target data \cite{DBLP:conf/cvpr/LiZXW14, DBLP:conf/iccv/ZhengSTWWT15, DBLP:conf/iccv/ZhengZY17, DBLP:conf/cvpr/WeiZ0018}.
In this paper, we call this line \textbf{target-only re-ID} \cite{lin2019aBottom}.
The second line is to exploit the labeled source data additionally.
For example, PUL \cite{DBLP:journals/tomccap/FanZYY18} first fine-tunes a deep CNN model on the labeled source dataset and then fine-tunes the model on the unlabeled target dataset in an unsupervised manner.
ECN \cite{DBLP:journals/corr/abs-1904-01990} fine-tunes the deep CNN model on the labeled source dataset and the unlabeled target dataset simultaneously.
In the community, the second line is usually referred to as \textbf{domain adaptive re-ID} \cite{DBLP:conf/cvpr/Deng0YK0J18, DBLP:conf/cvpr/WeiZ0018, DBLP:journals/tomccap/FanZYY18, DBLP:conf/cvpr/WangZGL18, DBLP:conf/bmvc/LinLLK18, DBLP:conf/eccv/ZhongZLY18, DBLP:journals/tip/ZhongZZLY19, DBLP:journals/corr/abs-1904-01990}.
We propose an Adaptive Exploration (AE) method to improve the deep re-ID model on the target domain.
We mainly discuss AE with the domain adaptive re-ID protocol.
However, it can also achieve comparable accuracy on the target-only re-ID.

To learn discriminative features in the target domain, AE is designed to maximize distances between all person images and minimize distances between similar person images in the feature space.
We leverage a non-parametric classifier, equipped with a feature memory, to achieve this objective.
The feature memory stores the features of all target person images.
After each iteration, the memory is updated according to the newly learned person image features.
The non-parametric classifier classifies person images according to its feature memory.
Specifically, to maximize distances between all person images, the classifier treats each person image as an individual class.
Given the feature and the index of a person image, by a softmax, the classifier tries to minimize the distance between the newly learned feature and its old feature in the memory, while maximizing the distances between the newly learned feature and other person image features in the memory.
To minimize distances between a group of similar person images, the classifier treats them as the same class.

However, it is challenging to select reliable similar samples in an unsupervised manner.
Especially in the early training stage, it is impossible to find a large number of reliable similar person images that are located near to each other in the feature space.
To alleviate this problem, the AE method adaptively selects a small amount of reliable similar samples at the beginning of the training, according to a similarity threshold.
When the model becomes stronger, more person images are adaptively selected as reliable samples, which in return benefits the training for the deep re-ID model.
In this way, the model is progressively improved.
With this adaptive selection, one problem is that the method may result in an unbalanced neighborhood distribution.
Once a person image has more neighborhoods than others, the sum of its losses will be larger, which force other images to move to it quickly. 
In consequence, images with more neighborhoods will make more images become their neighborhoods, which inevitably contains unreliable and noisy person images.
To alleviate this problem, we integrate a balance strategy into the method.
A penalty is employed into the loss function.
The penalty automatically changes the weight on the loss to make training balanced.

The contributions of this article are twofold:
\begin{itemize}
	\item We propose an AE method for unsupervised person re-ID. 
	The method maximizes distances of all target images and minimizes distances of similar target images.
	With a similarity threshold and a balance term, the method can adaptively find reliable similar target images.
	\item Extensive experiments on three large-scale datasets demonstrate the effectiveness of AE.
	Our method significantly improves the unsupervised person re-ID accuracy, including both target-only re-ID and domain adaptive re-ID.
\end{itemize}

\section{Related Work}

In this section, we briefly recount the previous work of person re-ID, including supervised person re-ID and unsupervised person re-ID.
In this paper, we separate unsupervised person re-ID into target-only re-ID and domain adaptive re-ID.

\subsection{Supervised Person Re-identification}
In recent years, deep learning has promoted great development in many computer vision tasks \cite{DBLP:journals/tip/SongZLGWH18,DBLP:journals/corr/abs-1708-02478}.
Compared with hand-crafted feature methods \cite{DBLP:conf/cvpr/FarenzenaBPMC10, DBLP:conf/cvpr/ZhaoOW13, DBLP:conf/cvpr/KostingerHWRB12, DBLP:journals/pami/ZhengGX13, DBLP:conf/cvpr/LiaoHZL15, DBLP:conf/iccv/ZhengSTWWT15, DBLP:journals/pr/WangWLGS18}, deep learning methods have dominated the person re-ID community.
Based on deep networks, supervised person re-ID has been extensively studied from general problems to specific problems.

For general CNN model, the siamese model \cite{DBLP:conf/icpr/YiLLL14, DBLP:conf/eccv/VariorSLXW16, DBLP:conf/eccv/VariorHW16, DBLP:conf/cvpr/LiZXW14, DBLP:conf/cvpr/ChengGZWZ16, DBLP:conf/eccv/ShiYZLLZL16} and the classification model \cite{DBLP:conf/cvpr/ZhengZSCYT17, DBLP:journals/tomccap/ZhengZY16} are studied.
Among the work of siamese model exploration, Yi \textit{et al.} \cite{DBLP:conf/icpr/YiLLL14} and Li \textit{et al.} \cite{DBLP:conf/cvpr/LiZXW14} firstly employed a siamese network for person re-ID and utilized part information in model training.
For the classification model, Zheng \textit{et al.} \cite{DBLP:conf/cvpr/ZhengZSCYT17} used a conventional fine-tuning approach, the ID-discriminative embedding (IDE).
Recently, Quan \textit{et al.} \cite{DBLP:conf/iccv/QuanDWZY19} used Neural Architecture Search (NAS) to search an efficient and effective CNN architecture for person re-ID.
In addition to the deep network, ranking loss \cite{DBLP:journals/tip/ChenGL16, DBLP:conf/cvpr/ChengGZWZ16, DBLP:conf/cvpr/ChenCZH17, DBLP:journals/corr/HermansBL17} and classification loss \cite{DBLP:conf/cvpr/LiZXW14, DBLP:conf/eccv/VariorHW16, DBLP:conf/cvpr/WangZLZZ16} are also explored.
Besides, to alleviate the over-fitting problem, several data augmentation methods have been proposed \cite{DBLP:journals/corr/abs-1708-04896, DBLP:conf/iccv/ZhengZY17, DBLP:journals/tip/ZhongZZLY19}. 
Among these data augmentation methods,
Zhong \textit{et al.} \cite{DBLP:journals/tip/ZhongZZLY19} changed camera styles for images as data augmentation to enhance the robustness of the model to camera variance.

Besides general problems, some specific problems of person re-ID are studied recently, 
such as pose variance problem \cite{DBLP:conf/iccv/SuLZX0T17}, viewpoint variance problem \cite{DBLP:journals/corr/abs-1812-02162}, background variance problem \cite{DBLP:conf/cvpr/TianYLLZSYW18} and occlusion problem \cite{DBLP:conf/iccv/MiaoWLDY19}.
To alleviate the pose variance problem, several methods \cite{DBLP:conf/cvpr/FarenzenaBPMC10, DBLP:journals/tip/ZhengHLY17, DBLP:conf/cvpr/ChoY16, DBLP:conf/iccv/SuLZX0T17, DBLP:conf/cvpr/SarfrazSES18} are proposed to learn pose invariant representations. Su \textit{et al.} \cite{DBLP:conf/iccv/SuLZX0T17} employed a pose-driven deep convolutional model to leverage the human part cues for person re-ID.
For viewpoint variance problem, some work \cite{DBLP:journals/pami/WuLR15, DBLP:conf/iccv/KaranamLR15, DBLP:conf/iccv/ZhengFLGYGW17, DBLP:journals/corr/abs-1812-02162} focus on it. 
Among them, Sun \textit{et al.} \cite{DBLP:journals/corr/abs-1812-02162} quantitatively analyzed and revealed the impacts of different viewpoints.
To alleviate the background variance problem, several methods \cite{ DBLP:conf/cvpr/Song0O018, DBLP:conf/cvpr/TianYLLZSYW18, DBLP:conf/cvpr/XiaoLWLW17} is designed to enhance the robustness of the re-ID model to background noises. Tian \textit{et al.} \cite{DBLP:conf/cvpr/TianYLLZSYW18} proposed a person-region guided pooling network and random background augmentation to alleviate background bias.

To alleviate domain bias, in this work, we focus on unsupervised person re-ID.

\subsection{Target-only Person Re-identification}

Target-only person re-ID \cite{DBLP:conf/cvpr/XiaoLWLW17, lin2019aBottom} only leverages unlabeled target data during training.
Although without the deep CNN, two hand-crafted feature methods \cite{DBLP:conf/cvpr/LiaoHZL15, DBLP:conf/iccv/ZhengSTWWT15} are also classified by us as target-only person re-ID methods.
It is because the hand-crafted feature can be directly used for the person re-ID without any training data.
In addition to hand-crafted feature methods, Xiao \textit{et al.} \cite{DBLP:conf/cvpr/XiaoLWLW17} introduced the feature memory with online instance matching (OIM) loss. These can be used in target-only person re-ID.
Lin \textit{et al.} \cite{lin2019aBottom} introduced bottom-up clustering (BUC) and achieved competitive accuracy with only unlabeled target data.

\subsection{Domain Adaptive Person Re-identification}
Domain adaptive person re-ID \cite{DBLP:conf/cvpr/Deng0YK0J18, DBLP:conf/cvpr/WeiZ0018, DBLP:journals/tomccap/FanZYY18, DBLP:conf/cvpr/WangZGL18, DBLP:conf/bmvc/LinLLK18, DBLP:conf/eccv/ZhongZLY18, DBLP:journals/tip/ZhongZZLY19, DBLP:journals/corr/abs-1904-01990, DBLP:conf/iccv/YuWZ17, DBLP:conf/cvpr/LiYLYDW18} exploits both labeled source data and unlabeled target data.
Among them, Yu \textit{et al.} \cite{DBLP:conf/iccv/YuWZ17} proposed clustering-based asymmetric metric learning called CAMEL.
Fan \textit{et al.} \cite{DBLP:journals/tomccap/FanZYY18} introduced a progressive algorithm with clustering and fine-tuning for re-ID model training.
Deng \textit{et al.} \cite{DBLP:conf/cvpr/Deng0YK0J18} and Wei \textit{et al.} \cite{DBLP:conf/cvpr/WeiZ0018} transferred source data to the target domain with CycleGAN \cite{DBLP:conf/iccv/ZhuPIE17} and employed the generated data to train the re-ID model.
Wang \textit{et al.} \cite{DBLP:conf/cvpr/WangZGL18} and Lin \textit{et al.} \cite{DBLP:conf/bmvc/LinLLK18} incorporated attribute information to enhance the scalability and usability of the re-ID model.
Zhong \textit{et al.} \cite{DBLP:journals/corr/abs-1904-01990} considered the intra-domain variation for the target domain and kept three underlying invariances during training, that is, exemplar-invariance, camera-invariance, and neighborhood-invariance.
When keeping neighborhood-invariance, they fixed the number of neighborhoods for each target image. 
However, this might be not reasonable during model training.
For example, in the early stage of training, only a few reliable neighborhoods can be found near each target image in the feature space.
In this case, fewer neighborhoods should be selected.
Besides, in the later stage of training, more reliable neighborhoods move close to the target images in the feature space.
In this case, more neighborhoods can be selected.
Therefore, in this work, an adaptive selection of neighborhoods is employed.
However, adaptive selection leads to different numbers of neighborhoods for different target images which may lead to an unbalanced training.
Thus, a balanced term is employed in our method.
With the adaptive selection and the balance term, the AE method manages to achieve competitive accuracy on both domain adaptive person re-ID and target-only person re-ID.

\begin{figure}[t]
\footnotesize
\centering
\subfigure[]
{
    \centering
    \includegraphics[width=0.98\textwidth]{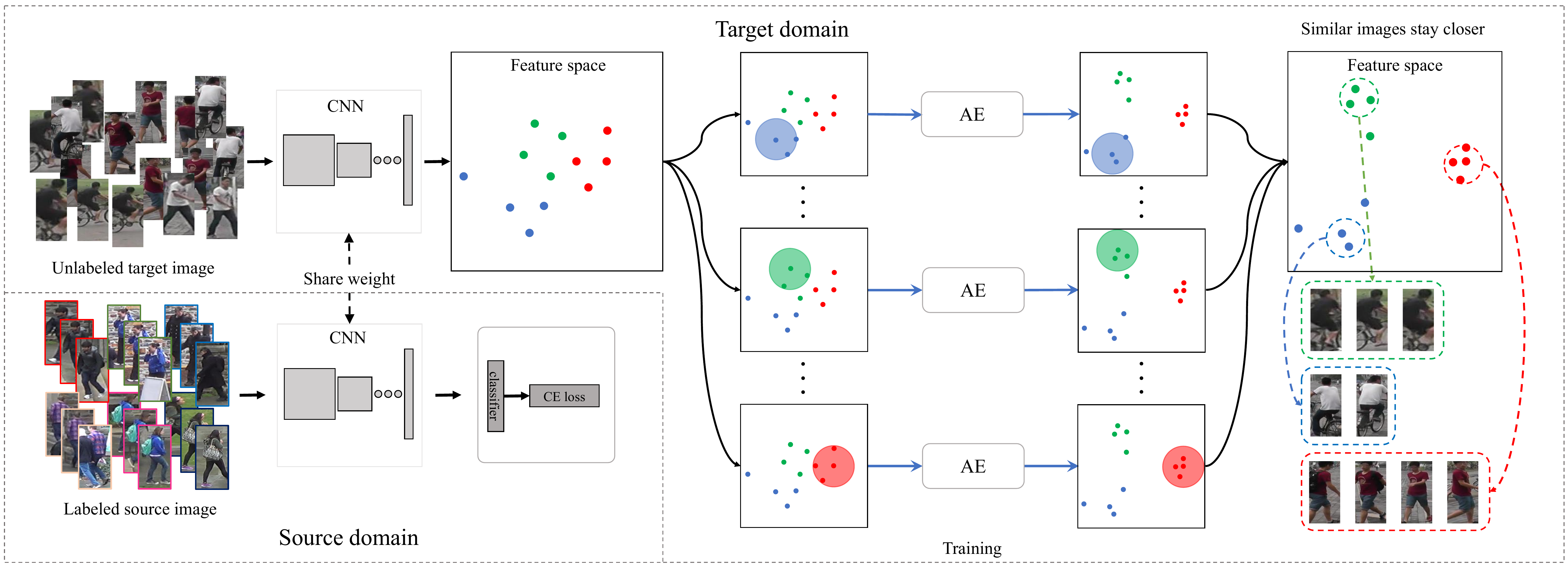}
    \label{Fig:framework}
}

\subfigure[]
{
    \centering
    \includegraphics[width=0.98\textwidth]{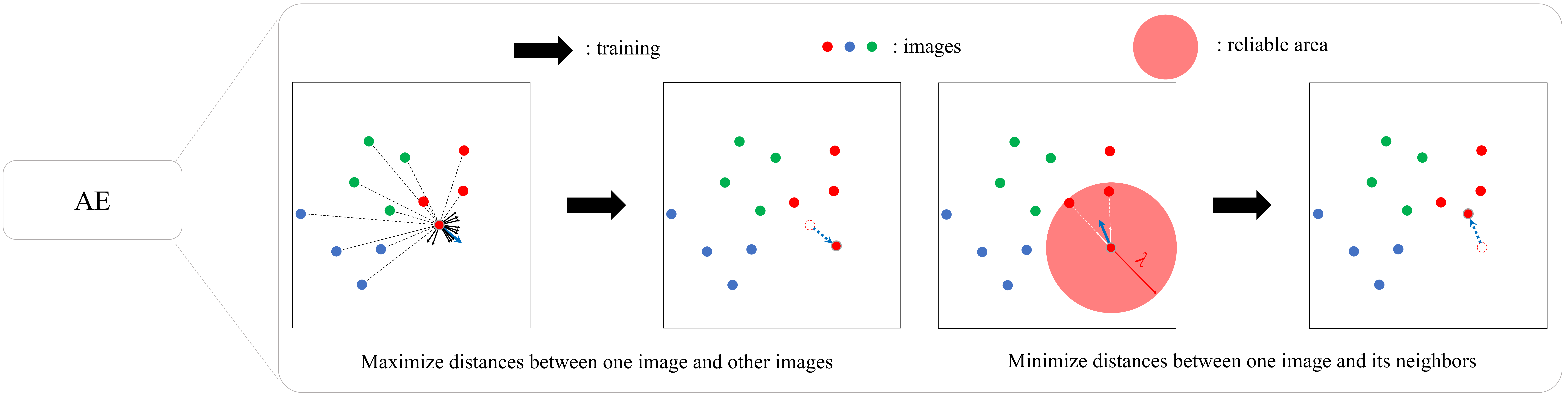}
    \label{Fig:distance}
}
\caption{(a) Overview of our framework for domain adaptive re-ID. 
The re-ID model is trained by source images and target images simultaneously. 
For source images, supervised learning is used. 
Specifically, a classifier with a cross-entropy (CE) loss is employed to classify persons in the source domain. 
For target images, the AE method is used. 
Specifically, AE aims to maximize feature distances between all images and minimize feature distances  between similar images. 
(b) Illustration of AE. 
The goal of AE is to help the re-ID model to learn discriminative features in the target domain. 
The basic idea is to encourage target images to keep away from each other while force similar target images to stay closer in the feature space.
When minimizing feature distances between similar images, according to a similarity threshold $\lambda$, only reliable samples are considered.
We employ a non-parametric classifier with a feature memory to train the re-ID model. Details about the non-parametric classifier are provided in Section~\ref{sec:Memory Network}.
}
\label{Fig:Frameworks}
\end{figure}

\section{Adaptive Exploration Method}{\label{sec:methods}}
In this section, we introduce the adaptive selection (AE) method in detail. 
The framework is shown in Figure~\ref{Fig:Frameworks}.
Source images and target images are input into a convolutional neural network (CNN) simultaneously.
The difference is that, with the labeled source images, supervised learning is used to train the model to obtain a basic re-ID capability. 
With the unlabeled target images, as shown in Figure~\ref{Fig:distance}, the AE method is proposed to help the model to generalize well in the target domain.
Using a non-parametric classifier equipped with a feature memory, our method simultaneously maximizes distances of all target images and minimizes distances of similar target images in the feature space.
In the end, our model can learn discriminative features for person images in the target domain.

In Section~\ref{sec:source domain}, we introduce supervised learning with parametric classifier in the source domain.
In Section~\ref{sec:Memory Network}, we introduce the non-parametric classifier, which is equipped with a feature memory and used for the target domain.
In Section~\ref{sec:Target domain}, we introduce the formulation of learning with adaptive selection, including maximizing distances between all images, minimizing distances between similar images, and adaptive selection of similar images.
In Section~\ref{sec:Diversity}, we introduce a balance strategy, which aims to make the number of neighborhoods balanced and reasonable.
In Section~\ref{sec:Overall Loss}, we introduce the optimization procedure of our method.

\subsection{Parametric Classifier for Source Domain}{\label{sec:source domain}}
In the source domain, supervised learning with parametric classifier is used.
Suppose the source domain contains $N_s$ labeled images $\{(x^s_1, y^s_1), (x^s_2, y^s_2), ..., (x^s_{N_s}, y^s_{N_s})\}$, where $s$ denotes the source domain.
The probability that the $x^s_i$ image belongs to the $y^s_i$ identity is defined as follows, 

\begin{equation}\label{eq:parametric classifier}
p(y_i^s|x_i^s) = \frac{\mathrm{exp}(g(\phi(x_i^s;\boldsymbol{\theta}); \mathbf{w})[y_i^s])}
{\sum_{j=1}^J \mathrm{exp}(g(\phi(x_i^s;\boldsymbol{\theta}); \mathbf{w})[j])},
\end{equation}
where $\phi(\cdot ;\boldsymbol{\theta})$ denotes our model, extracting features for every image, $\boldsymbol{\theta}$ denotes weights of the deep re-ID model,
$g(\cdot; \mathbf{w})$ denotes the parametric classifier, $\mathbf{w}$ denotes the weights of the classifier,
$g( \phi(x_i^s;\boldsymbol{\theta}); \mathbf{w})[j]$ denotes the $j$-th logit of the output from the classifier given the feature $\phi(x_i^s;\boldsymbol{\theta})$ and $J$ denotes the number of person identities of the source dataset.
The training for the source domain is to minimize the following loss,
\begin{equation}\label{eq:source domain}
L_s = - \frac{1}{N_s} \sum_{i=1}^{N_s} \mathrm{log} p(y_i^s|x_i^s).
\end{equation}

\subsection{Non-parametric Classifier with Feature Memory}{\label{sec:Memory Network}}
To learn discriminative features for the target domain, our method tries to maximize feature distances between all person images and minimize feature distances between similar person images.
To optimize distances, one can use contrastive loss \cite{DBLP:conf/eccv/VariorHW16} or triplet loss \cite{DBLP:journals/corr/HermansBL17, DBLP:journals/tip/LiuFQJY17}.
However, these losses become less effective when datasets become large.
Therefore, we try to optimize distances under a classification framework.
For example, to minimize feature distances between similar person images, we can treat them as the same class. 
To maximize feature distances between all target person images, we can treat each image as an individual class.

However, treating each image as an individual class may make the parametric classifier difficult to converge.
To alleviate this problem, motivated by~\cite{DBLP:conf/cvpr/WuXYL18,DBLP:conf/eccv/WuEY18,DBLP:conf/cvpr/XiaoLWLW17,DBLP:journals/corr/abs-1904-01990,yang2019patch}, we exploit a non-parametric classifier, which is equipped with a memory \textbf{M}, to classify target images.

Suppose the target domain contains $N_t$ unlabeled images $\{x^t_1, x^t_2, ..., x^t_{N_t}\}$, where $t$ denotes the target domain.
After feature extraction, each image is embedded as a $D$-dimensional vector.
The feature memory $\textbf{M} \in \mathbb{R}^{D \times N_t}$ stores all target image features and is updated after each training iteration.

Based on the memory $\textbf{M}$, given an image, the non-parametric classifier aims to produce the probability of the image being the same class as other images. For example, the probability that the $x_i^t$ image is same as the $k$-th image is defined as follows,

\begin{equation}\label{eq:non-parametric classifier}
p(k|x_i^t) = \frac{\mathrm{exp}(\textbf{M}[k]^T \phi(x_i^t;\boldsymbol{\theta})/\tau)}{\sum_{j=1}^{N_t} \mathrm{exp}(\textbf{M}[j]^T \phi(x_i^t; \boldsymbol{\theta})/ \tau)},
\end{equation}
where $\textbf{M}[k]$ denotes the $k$-th column of the feature memory \textbf{M}, representing the feature of the $k$-th image,
and the hyper-parameter $\tau$ denotes the temperature fact of the softmax function. 
A higher temperature $\tau$ leads to a softer probability distribution. 
After each iteration, $\textbf{M}$ is updated as follows, 
\begin{equation}\label{eq:memory update}
\textbf{M}[i] \leftarrow \mu\textbf{M}[i] + (1 - \mu)\phi(x_i^t;\boldsymbol{\theta}),
\end{equation}
where the hyper-parameter $\mu$ is the update rate of \textbf{M}. 

Instead of fixing $\mu$ to a constant value, we increase it linearly as the number of epochs increasing.
Since \textbf{M} is not reliable enough at the beginning of training, a smaller $\mu$ is needed to accelerate the update of \textbf{M}.
By rapidly updated with newly learned representations, \textbf{M} can memorize discriminative features quickly.
As \textbf{M} becomes discriminative gradually, \textbf{M} is required to be more stable.
Therefore, in this time, a larger $\mu$ is used to slow down the update.

\subsection{Learning with Adaptive Selection}{\label{sec:Target domain}}
In this subsection, we introduce the learning with adaptive selection for the target domain, including maximizing distances between all target images, minimizing distances between similar target images, and adaptive selection of similar images.

We firstly introduce maximizing distances between all target images.
To achieve this objective, we assume each target image as an individual class.
Specifically, the index of each target image is treated as its pseudo label. 
We try to maximize the distances between target images by minimizing the following loss,

\begin{equation}\label{eq:maximize distances}
L_{\alpha} = -\frac{1}{N_t} \sum_{i=1}^{N_t} \mathrm{log}p(i|x_i^t).
\end{equation}
When applying this loss to a specific image, the image is encouraged to move far away from other images (Figure~\ref{Fig:distance}).
When applying this loss to all images, they are encouraged to move far away from each other.
The feature distances between all target images are thus maximized.

To minimize distances between similar target images,
we first use a similarity threshold to adaptively select reliable neighborhoods for each target image. 
Only neighborhoods whose distances to the given target image are smaller than the threshold are selected as reliable neighborhoods.
Secondly, we assume that the target image and its reliable neighborhoods share the same person identity, \textit{i.e.}, treating the image and its reliable neighborhoods as the same class.
By this operation, each target image is forced by the non-parametric classifier to move closer to its neighborhoods (Figure~\ref{Fig:distance}), which makes similar target images stay closer.
The loss is defined as follows,
\begin{equation}\label{eq:minimize distances wologk}
L_{\beta1} = -\frac{1}{N_t} \sum_{i=1}^{N_t}\sum_{j=1, j \neq i}^{N_t}  \mathbf{v}_i^j \mathrm{log}p(j|x_i^t),
\end{equation}
where $\mathbf{v}_i \in \{0, 1\}^{N_t}$ denotes the selection indication vector of $i$-th target image. 
Specifically, $\mathbf{v}_i^j = 1$ indicates that the $j$-th image is selected as a neighborhood of the $i$-th image.
When $\mathbf{v}_i^j = 0$, the $j$-th image will not be used when forcing the $i$-th image move closer to its neighborhoods.

For neighborhoods selection, we are inspired by active learning \cite{DBLP:journals/tmm/YanNLGYX16,DBLP:journals/tgrs/DengXLLT19,DBLP:journals/pr/DengLLT18} and Self-Pace Learning (SPL) \cite{DBLP:conf/nips/KumarPK10}, which has been widely used in weakly-supervised learning~\cite{DBLP:conf/iccv/FanCCYXH17}, semi-supervised learning~\cite{DBLP:journals/tomccap/FanZYY18,  DBLP:conf/icml/MaMXLD17} and unsupervised learning~\cite{DBLP:journals/tomccap/FanZYY18}. 
The basic idea of SPL is to incorporate from easy samples to hard samples during training.
Samples with small losses are considered as reliable samples.
In this paper, reliable or easy neighborhoods are needed to be selected for minimizing feature distances between similar person images.
We select reliable neighborhoods according to their distances or similarities to the given image.
Specifically, when an image is close enough to the given image in the feature space, it is selected as a reliable neighborhood to the given image.
We formulate this selection as minimizing the following loss, 

\begin{equation}\label{eq:neighborhoods selection 0}
\begin{split}
\begin{aligned}
L_{\gamma} = \sum_{i=1}^{N_t} \sum_{j=1}^{N_t} \mathbf{v}_i^j & \mathrm{d}(\textbf{M}[j], \phi(x_i^t;\boldsymbol{\theta})) -  \lambda \Vert \mathbf{v}_i \Vert_1, \\
& s.t. \: \mathbf{v}_i^j \in \{0,1\}
\end{aligned}
\end{split}
\end{equation}
where $\lambda$ is the similarity threshold, $\mathrm{d}(\cdot)$ denotes a distance function and $\Vert \mathbf{v}_i \Vert_1$ denotes the number of selected neighborhoods for the $i$-th image.
By minimizing $L_{\gamma}$, two images will be treated as neighborhoods if their distance is less than $\lambda$.

\begin{figure}[h]
\footnotesize
\centering
\subfigure[without balance]
{
    \centering
    \includegraphics[width=0.45\textwidth]{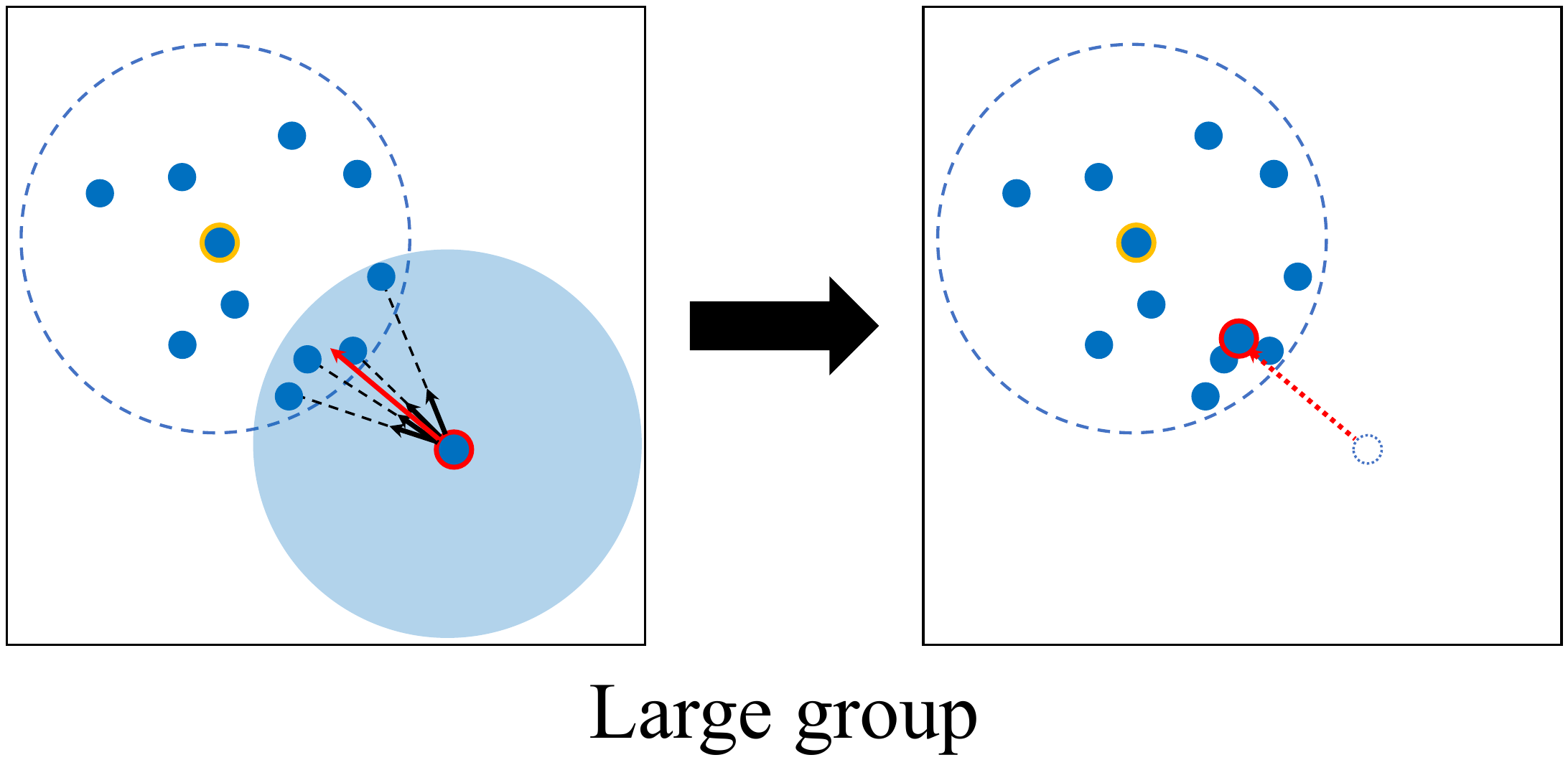}\hspace{10pt}
    \includegraphics[width=0.45\textwidth]{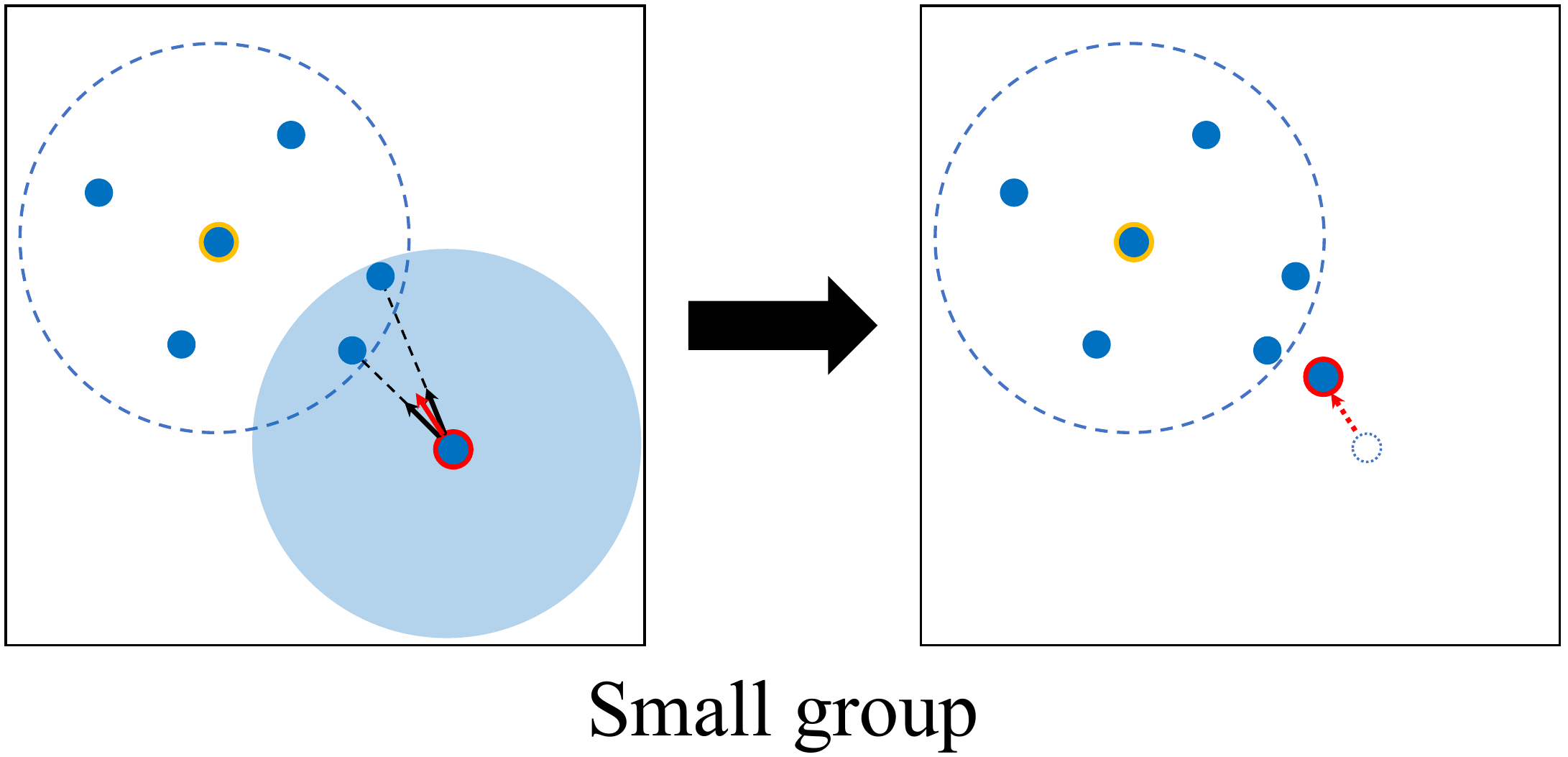}
    \label{Fig:without balance illustration}
}

\subfigure[with balance]
{
    \centering
    \includegraphics[width=0.45\textwidth]{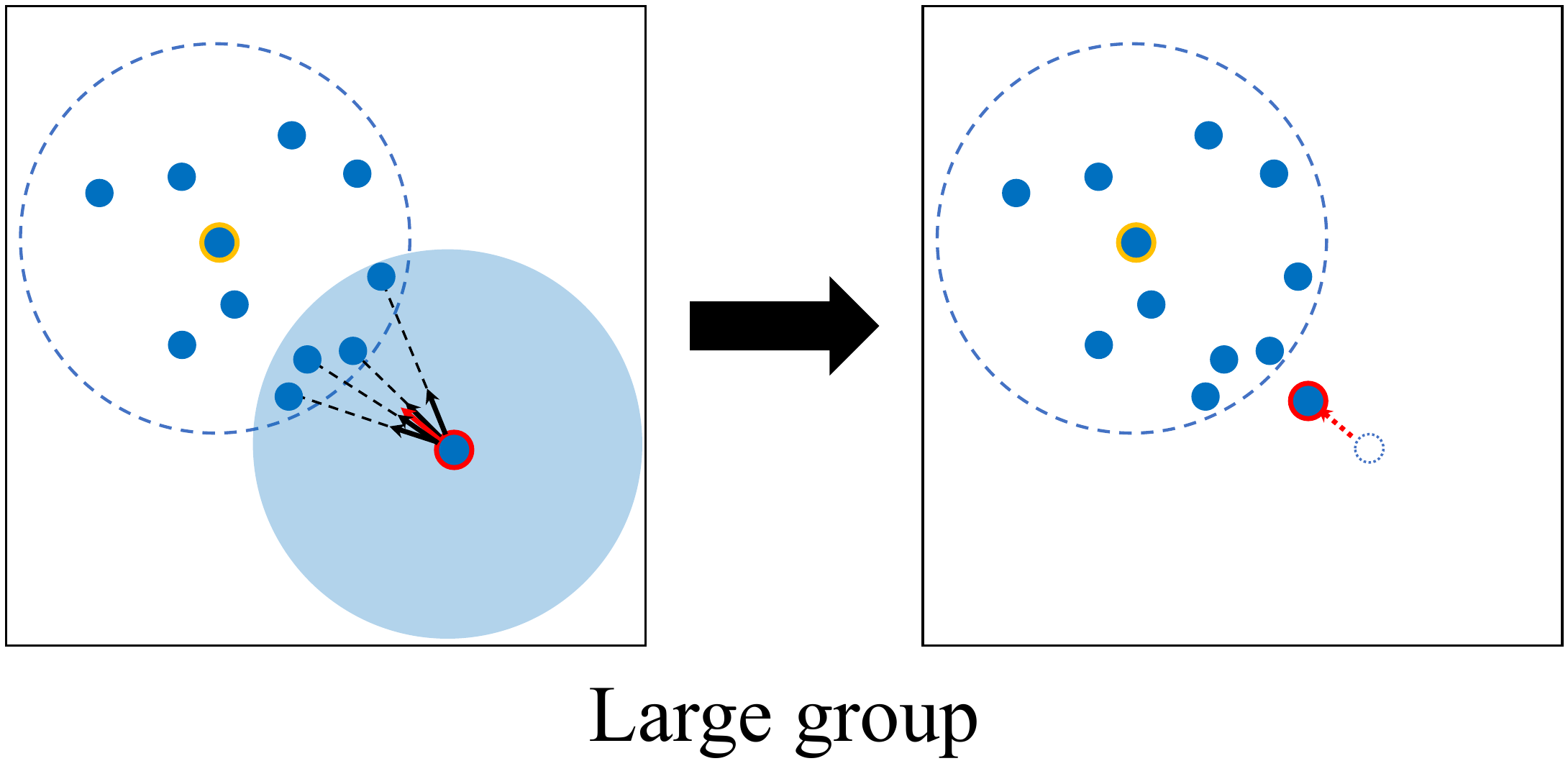}\hspace{10pt}
    \includegraphics[width=0.45\textwidth]{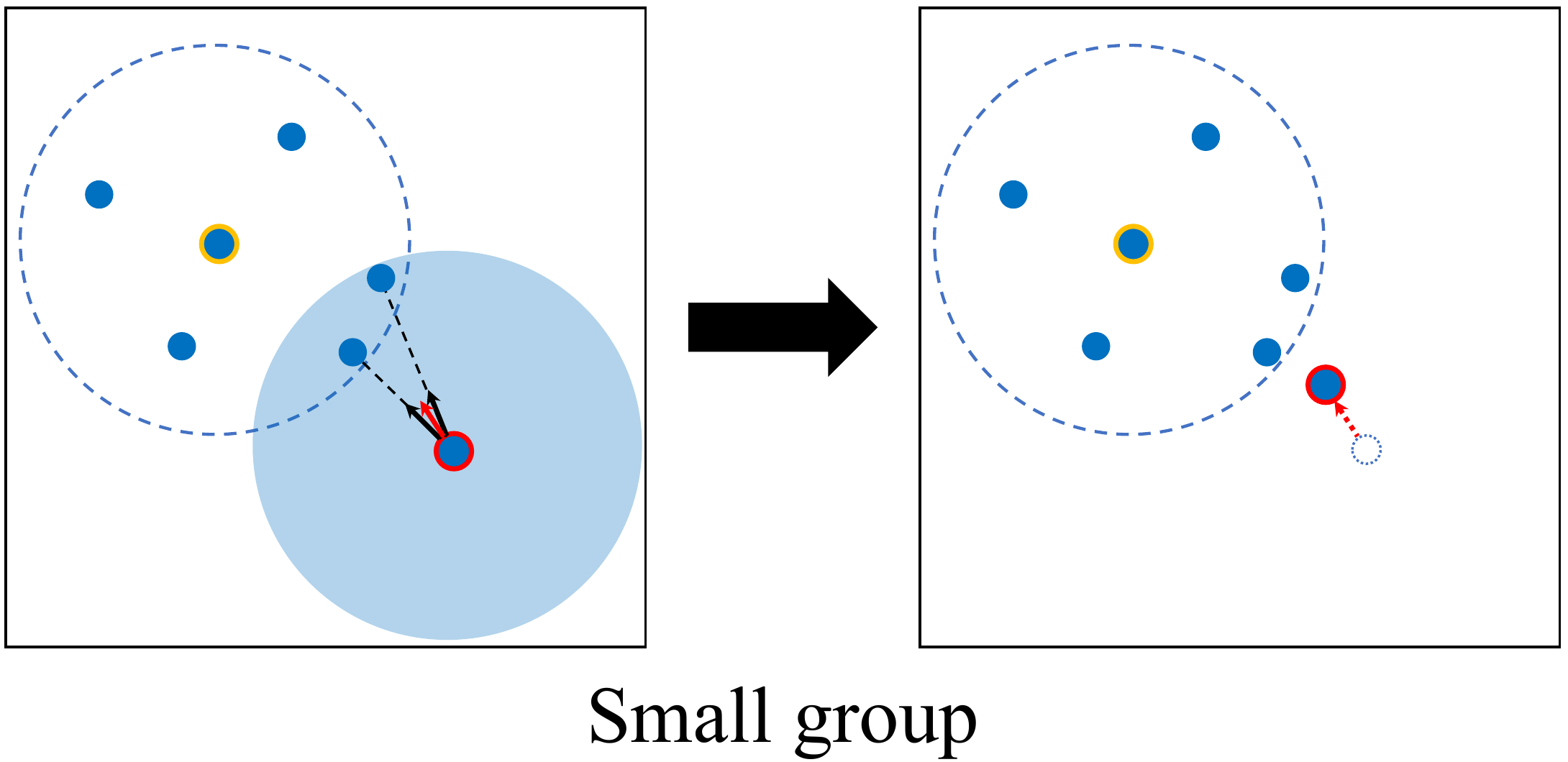}
    \label{Fig:with balance illustration}
}

\caption{Illustration of learning with balance and without balance.
(a) Learning without balance. When an image (the point with yellow border) has a large number of neighborhoods (points in the dashed circle), because some of its neighborhoods (points in the intersection of the dashed circle and the blue circle) can also be the neighborhoods of another image (the point with red border), it is easier for a large group to attract other images than a small group.
(b) Learning with balance. When an image has a large number of neighborhoods, the balance strategy decreases the losses between the image and its neighborhoods (points in the intersection of the dashed circle and the blue circle) by a larger penalty. Otherwise, the losses are decreased by a smaller penalty.
As a result, no matter a large group or a small group, they attract images in a relatively similar degree, which makes the number of neighborhoods balanced and reasonable. 
}
\label{Fig:illustration for balance}
\end{figure}

\subsection{Learning with Balance}{\label{sec:Diversity}}
A problem caused by Eq.~(\ref{eq:minimize distances wologk}) is that the number of neighborhoods for an image can change dramatically.
When an image has a large number of neighborhoods, the sum of the losses between it and its neighborhoods can be considerably large.
When an image has a small number of neighborhoods, the sum of the losses between it and its neighborhoods can be very small.
As a consequence, as shown in Figure~\ref{Fig:without balance illustration}, it is easier for a large group to attract other images than a small group, which makes most groups have only a few data instances while a minority of groups have a large number of data instances.
This unbalanced learning may result in a poor accuracy for re-ID.

To address this issue, we integrate a balance term into Eq.~(\ref{eq:minimize distances wologk}) to make the adaptive selection balanced,
\begin{equation}\label{eq:minimize distances wilogk}
L_{\beta2} = -\frac{1}{N_t} \sum_{i=1, \Vert \mathbf{v}_i {\Vert}_1 \geq 2 }^{N_t}
\frac{1}{\Vert \mathbf{v}_i {\Vert}_1 \mathrm{log}(\Vert \mathbf{v}_i {\Vert}_1)} \sum_{j=1, j \neq i}^{N_t} \mathbf{v}_i^j \mathrm{log}p(j|x_i^t).
\end{equation}
When $\Vert \mathbf{v}_i {\Vert}_1 = 1$, the image does not have any other  neighborhoods expect itself.
Therefore, we don't use it as a training sample and only consider the cases when $\Vert \mathbf{v}_i {\Vert}_1 \geq 2$.

When an image has a large number of  neighborhoods, the penalty $1/(\Vert \mathbf{v}_i \Vert_1 \mathrm{log}(\Vert \mathbf{v}_i \Vert_1))$ decreases the losses between the image and its neighborhoods heavily. Otherwise, the losses are decreased slightly (Figure~\ref{Fig:with balance illustration}).
By this balance strategy, no matter whether it's a large group or a small group, the group attracts images in a relatively similar degree.
Therefore, they will have a similar number of neighborhoods.
During training, the number of neighborhoods thus becomes balanced and reasonable.




\begin{algorithm}[t]
\SetAlgoLined
\SetKwInOut{Input}{Input}
\SetKwInOut{Output}{Output}
\SetKwInOut{Initialization}{Initialization}
\caption{Adaptive Exploration for Domain Adaptive Re-ID}
\label{alg: Framework}
\Input{Unlabeled target data $\{x_i^t\}_{i=1}^{N_t}$; \\
       Labeled source data $\{x_i^s, y_i^s\}_{i=1}^{N_s}$; \\
       Similarity threshold $\lambda$; \\
       Update rate of the feature memory $\mu$; \\
       Number of epochs $K$; \\
       Original model $\phi(\cdot;\boldsymbol{\theta}_0)$.
       }
\Output{Model $\phi(\cdot; \boldsymbol{\theta}_K)$.
       }
\Initialization{randomly initialize $\mathbf{w}_0$; \\
                zero initialize the feature memory \textbf{M}.}
\For{$k = 0$ to $K-1$}{\label{alg:epoch section}
    // adaptive selection \label{alg: neighborhoods sample start} \\
    \For{$i = 1$ to $N_t$}{ \label{alg:target section} 
        extract feature $\mathbf{f}_i$ =\  $\phi(x_i^t; \boldsymbol{\theta}_k)$\;
        calculate distances: $ \mathrm{d}(\textbf{M}, \mathbf{f}_i)$; \\
        \For{$j = 1$ to $N_t$}{
            \eIf{$ \mathrm{d}(\mathrm{\textbf{M}[j]}, \mathbf{f}_i) <= \lambda$ }{
                $\mathbf{v}_i^j = 1$; // selected 
            }{
                $\mathbf{v}_i^j = 0$; // removed
            }
        }
    }\label{alg: neighborhoods sample end}
    // model training \\
    train the model $<\phi(\cdot; \boldsymbol{\theta}_k), \mathbf{w}_k>$ with source images and selected target images:
    $\boldsymbol{\theta}_{k} \rightarrow \boldsymbol{\theta}_{k+1}$, $\mathbf{w}_{k} \rightarrow \mathbf{w}_{k+1}$; \\
    // feature memory \textbf{M} update \\
    \For{$i = 1$ to $N_t$ }{\label{alg: update features}
        $\textbf{M}[i] \leftarrow \mu \textbf{M}[i] + (1 - \mu) \mathbf{f}_i $;
    }
}
\end{algorithm}

\subsection{Optimization Procedure}{\label{sec:Overall Loss}}

During training, we alternately optimize the involved parameters, \textit{i.e.}, \textbf{v} and ($\boldsymbol{\theta}, \mathbf{w}$).

\textit{1) Optimize $\mathbf{v}$ when $(\boldsymbol{\theta}, \mathbf{w})$ is fixed.} 
The goal of this step is to adaptively select reliable neighborhoods for minimizing distances between them in the feature space,
which is achieved by minimizing

\begin{equation}\label{eq:neighborhoods selection}
\underset{\mathbf{v}} {\mathrm{min}} \ \ \ L_{\gamma}.
\end{equation}
Specially, if the distance between two images is below the threshold $\lambda$, they will be chosen as neighborhoods for each other. The details are provided in Algorithm~\ref{alg: Framework}.

\textit{2) Optimize $(\boldsymbol{\theta}, \mathbf{w})$ when $\mathbf{v}$ is fixed.} This step utilizes source data and target data to train the re-ID model by minimizing
\begin{equation}\label{eq:model training}
\underset{\mathbf{\boldsymbol{\theta}, \mathbf{w}}} {\mathrm{min}} \ \ \xi(L_{\alpha} + \delta L_{\beta2}) + (1 - \xi) L_s,
\end{equation}
where hyper-parameter $\delta$ and $\xi$ aim to control the importance of these losses.

\textit{3) Update the feature memory $\mathbf{M}$.} In this step, $\mathbf{M}$ for the non-parametric classifier is updated by Eq.~(\ref{eq:memory update}).

With the optimization procedure, our model manages to recognize people in the target domain. 
Note that, with the target-only re-ID protocol, the loss from source domain is not used and thus the $\xi$ in Eq.~(\ref{eq:model training}) is set to 1.

\section{Experiment}

In this section, we evaluate the proposed method on three large-scale re-ID datasets.
Besides Market-1501 and DukeMTMC-reID, which are widely used by most existing methods, we also report the accuracy on the MSMT17 dataset.

\subsection{Datasets and Settings}

\textbf{Market-1501} \cite{DBLP:conf/iccv/ZhengSTWWT15} contains 32,688 images with 1,501 identities.
They are captured by 6 cameras on campus.
The dataset is split into three parts: 12,936 images with 751 identities for training, 19,732 images with 750 identities for the gallery, and another 3,368
images with the same 750 gallery identities for query.

\textbf{DukeMTMC-reID} \cite{DBLP:conf/iccv/ZhengZY17} contains 36,411 images with 1,812 identities, which are collected from 8 cameras. 
Following Market-1501, the dataset is split into three parts: 16,522 images with 702 identities for training,
17,661 images with 1,110 identities in the gallery, and another 2,228 images with the same 702 identities as the
gallery for query.

\textbf{MSMT17} \cite{DBLP:conf/cvpr/WeiZ0018} contains 126,441 images with 4,101 identities, which are captured by 15 cameras.
Similar to Market-1501 and DukeMTMC-reID, the dataset is split into three parts: 32,621 images with 1,041 identities for training,
82,161 images with 3,060 identities in the gallery and another 11,659 images with the same 3,060 gallery identities for query.

We reported the rank1, rank5, rank10 and mean average precision (mAP) for evaluation on the three datasets.
All experiments used single-query. 

\subsection{Implementation Details}{\label{sec:implementation}}
We used ResNet-50 \cite{DBLP:conf/cvpr/HeZRS16} as the CNN backbone to extract features. 
The ResNet-50 was pre-trained on ImageNet \cite{DBLP:conf/cvpr/DengDSLL009}.
After the Pool-5 layer of the ResNet-50, we added a 4,096-dimension fully-connected layer followed by batch normalization \cite{DBLP:conf/icml/IoffeS15}, ReLU \cite{DBLP:conf/icml/NairH10} and Dropout \cite{DBLP:journals/jmlr/SrivastavaHKSS14}.
Therefore, the length of re-ID features for training is 4,096.
Note that, during testing, following most existing methods, we used the 2,048-dimension feature from the Pool-5 layer.
During evaluation, the 2048-dimension feature is able to achieve similar accuracy as the 4096-dimension feature, but consumes fewer flops and parameters .
During training, we fixed the first two layers of the ResNet-50.
After feature extraction, two different classifiers were used to classify person images.
For source images, we adopted a general parametric classifier for supervised learning.
For target images, we adopted a non-parametric classifier with a feature memory \textbf{M}.

We used random flip, crop and erasing \cite{DBLP:journals/corr/abs-1708-04896} as data augmentation for both target images and source images.
Images were resized to 256 $\times$ 128.
For each iteration, we randomly chose 128 images from the target domain and 128 images from the source domain.
We also leveraged the CamStyle \cite{DBLP:journals/tip/ZhongZZLY19} method to reduce camera variance in the target domain.

We used Stochastic Gradient Descent (SGD) to train the model.
Weight decay and momentum were set to $5 \times 10^{-4}$ and 0.9, respectively. 
Learning rate  was set to 0.01 for ResNet-50 base layers and 0.1 for other layers in the first 40 epochs. 
The learning rate was then divided by 10 in the next 20 epochs.
For the target domain, we started minimizing distances of similar images after 5 epochs.
During training, we used cosine to measure similarity.
The 4,096-dimension features are $l_2$-normalized before stored into the memory \textbf{M}.
Two images will be selected as neighborhoods when their cosine similarity is larger than $\lambda$.

For hyper-parameters, we set the similarity threshold $\lambda$ and the temperature $\tau$ to 0.55 and 0.05, respectively.
By default, with the number of epochs increasing, the memory update rate $\mu$ was linearly increased from 0 to 0.4 for Market-1501 and from 0 to 0.5 for DukeMTMC-reID.
The $\delta$ and $\xi$ was set to 3.5 and 0.6, respectively. 

\begin{figure*}[h]
\centering
\subfigure[DukeMTMC-reID to Market-1501]
{
    \centering
    \includegraphics[width=0.315\textwidth]{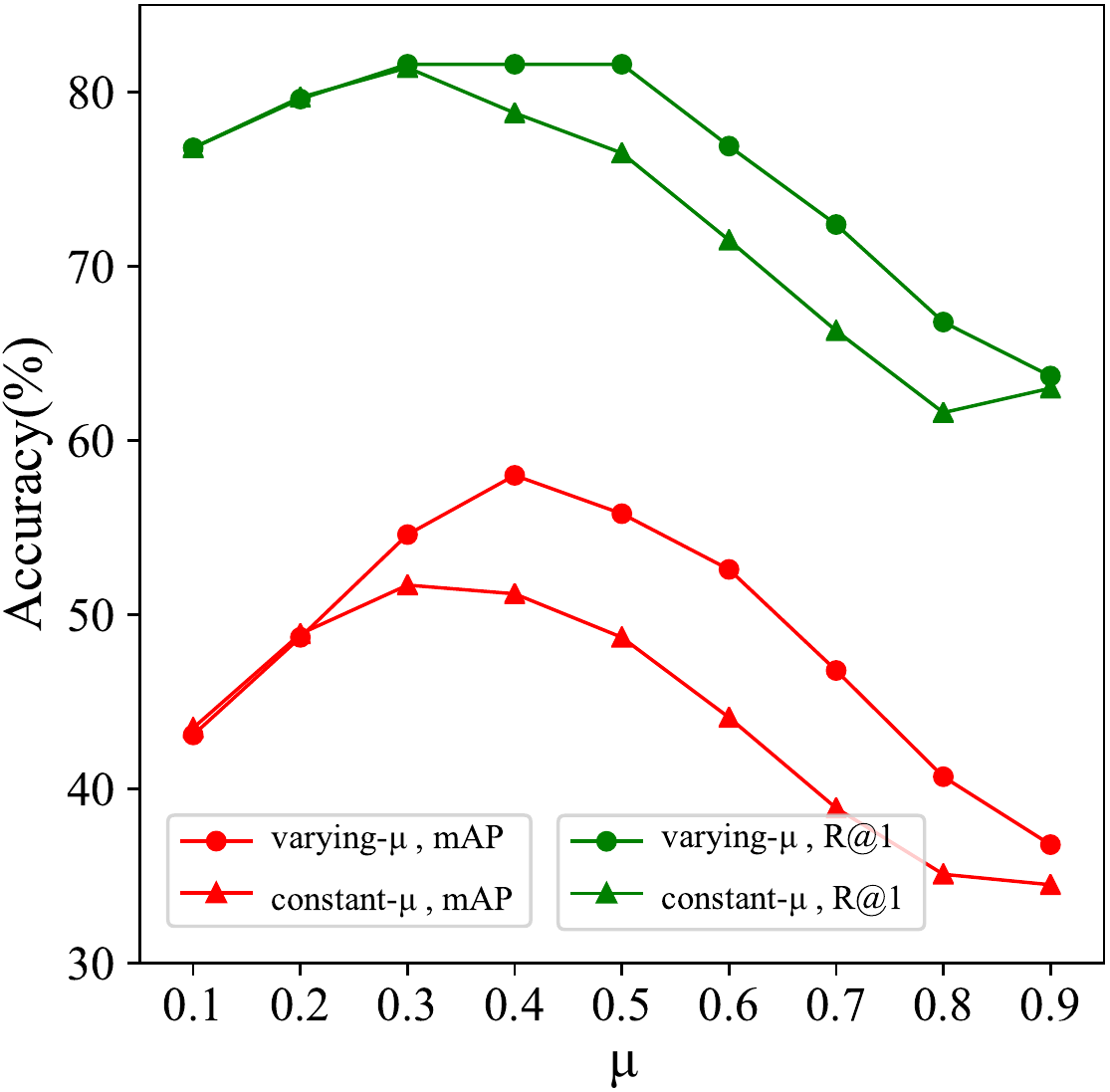}
    \label{Fig:mu_d2m}
}
\subfigure[Market-1501 to DukeMTMC-reID]
{
    \centering
    \includegraphics[width=0.315\textwidth]{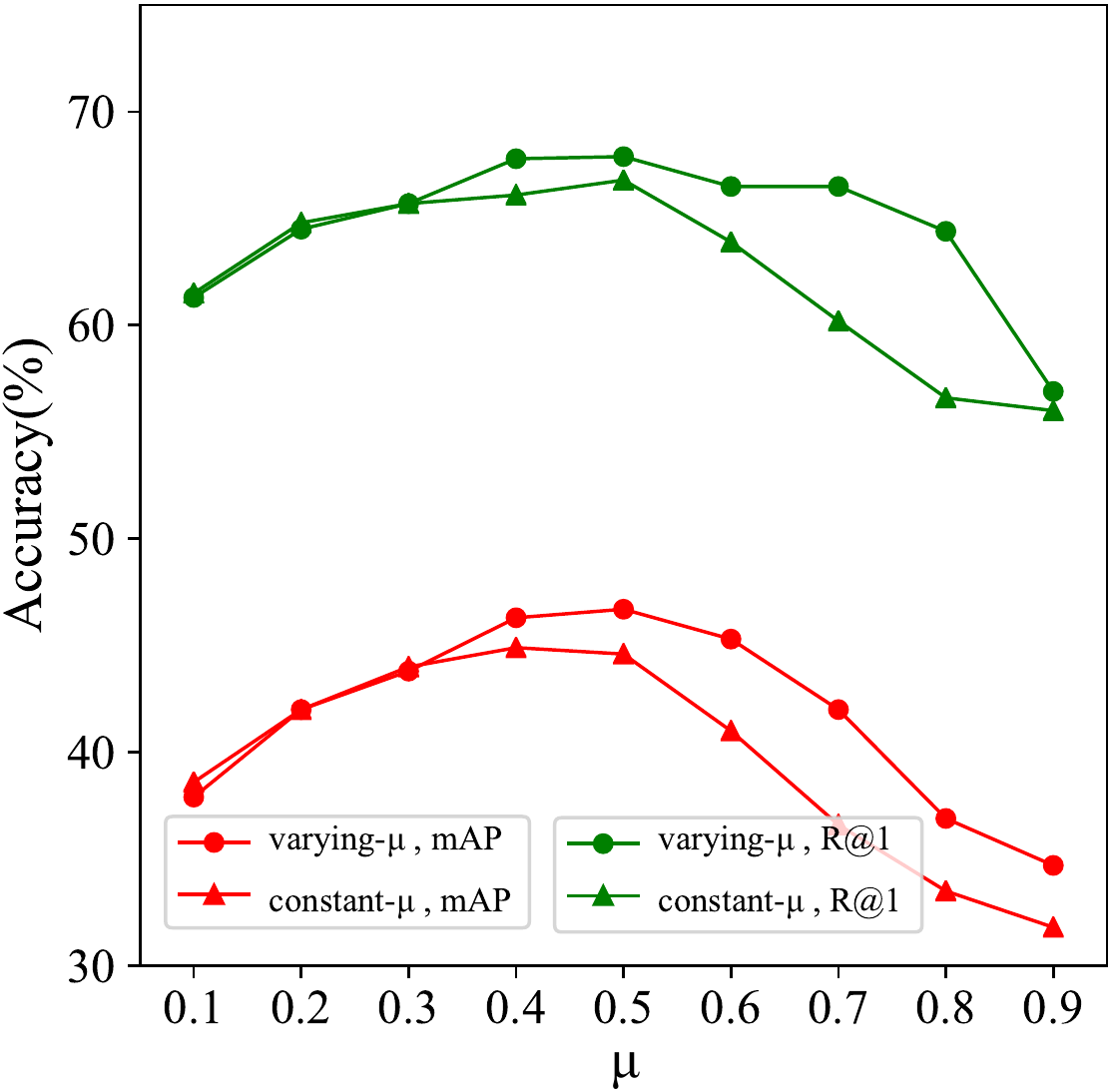}
    \label{Fig:mu_m2d}
}
\subfigure[Similarity Threshold]
{
    \centering
    \includegraphics[width=0.315\textwidth]{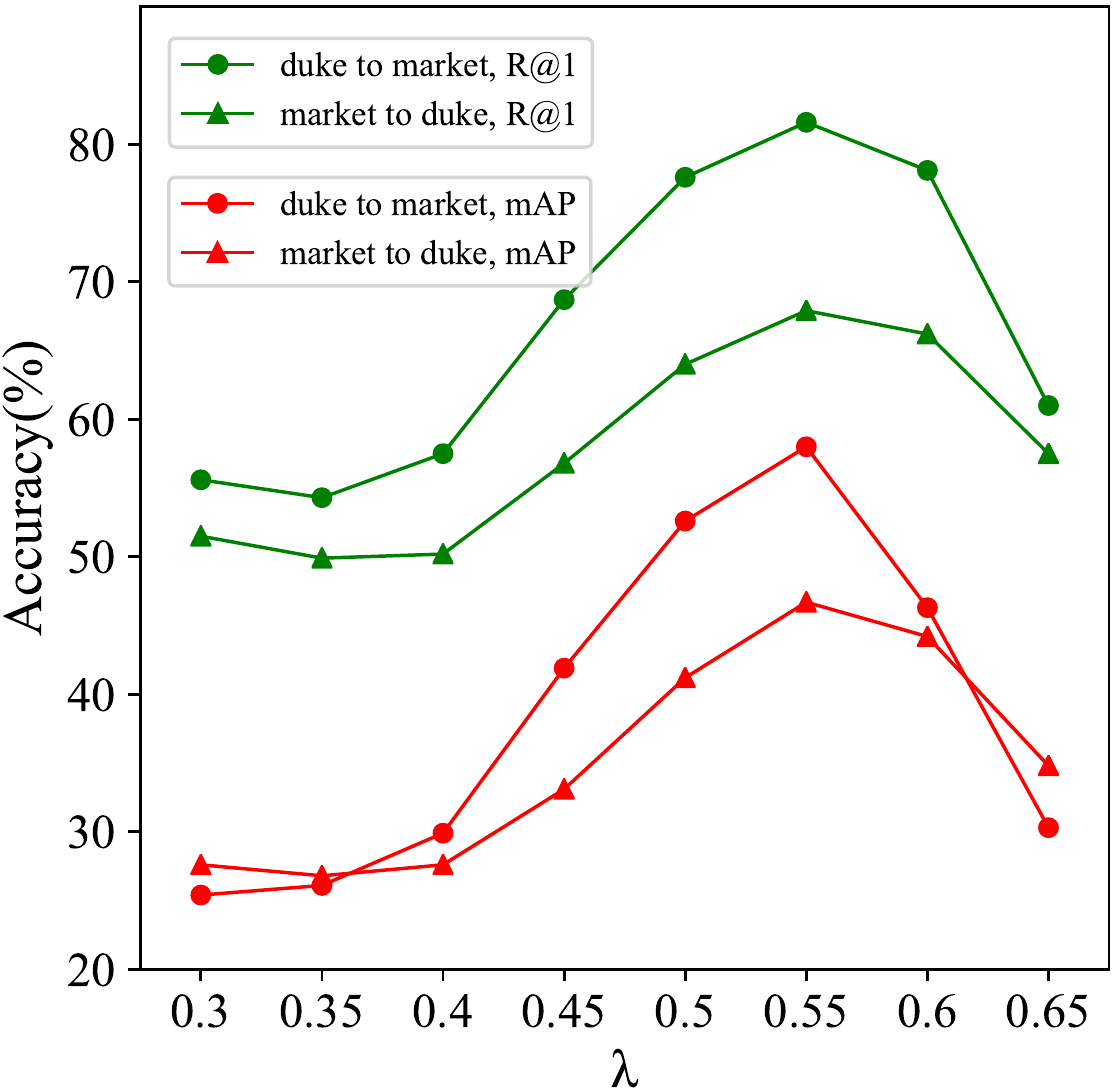}
    \label{Fig:lambda_change}
}

\subfigure[Similarity Threshold]
{
    \centering
    \includegraphics[width=0.315\textwidth]{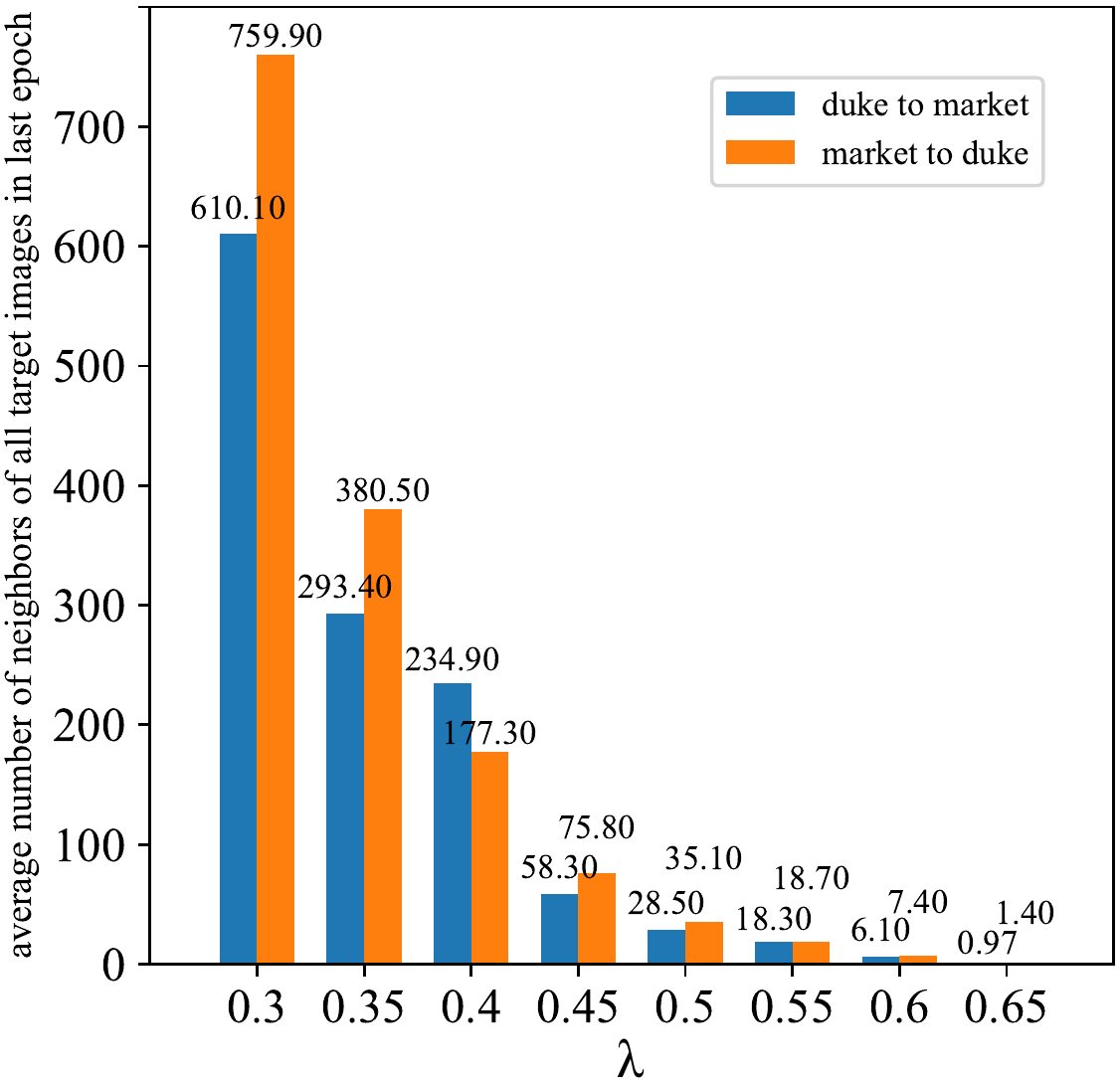}
    \label{Fig:last_kmean}
}
\subfigure[DukeMTMC-reID to Market-1501]
{
    \centering
    \includegraphics[width=0.315\textwidth]{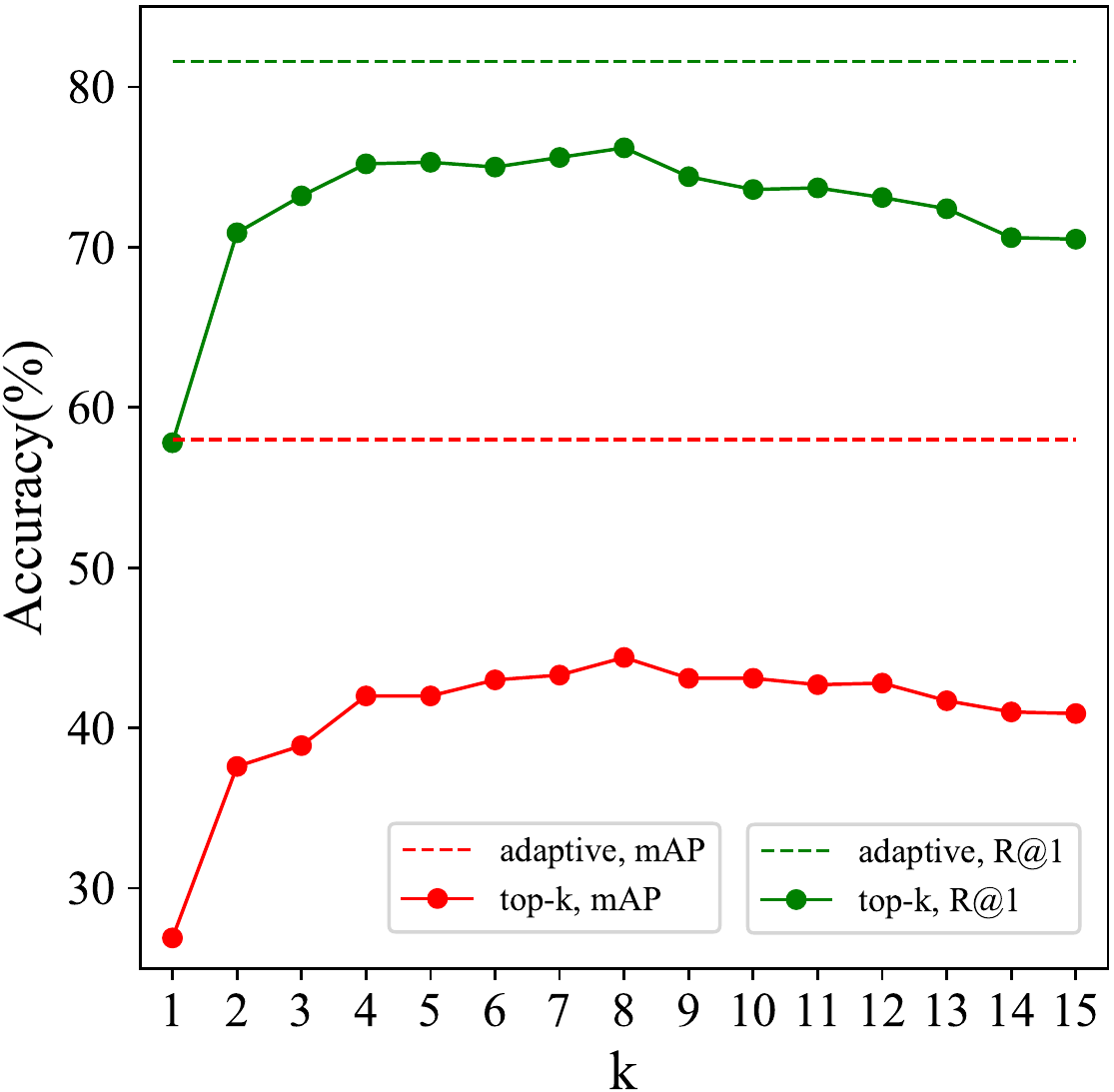}
    \label{Fig:k_d2m}
}
\subfigure[Market-1501 to DukeMTMC-reID]
{
    \centering
    \includegraphics[width=0.315\textwidth]{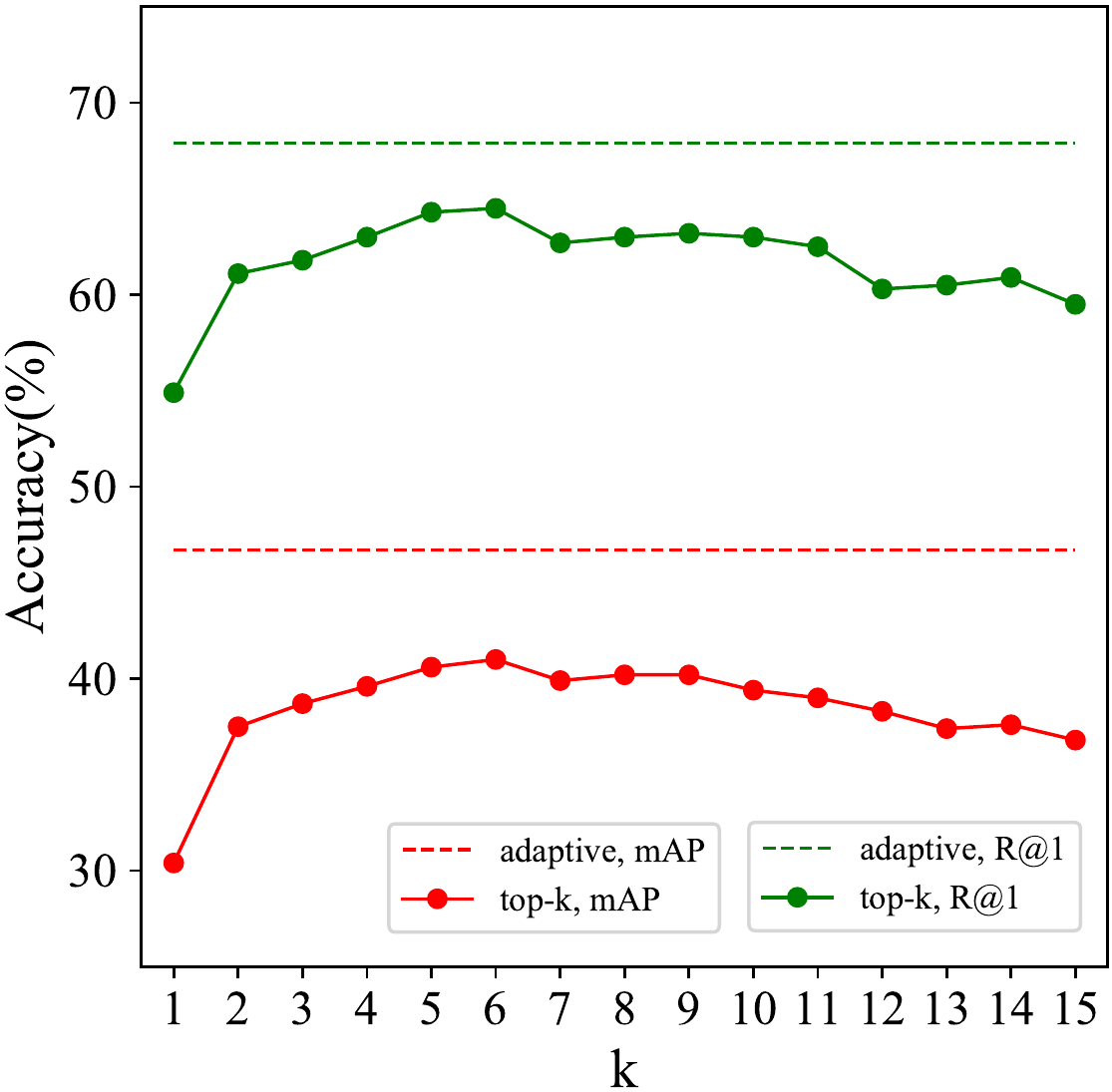}
    \label{Fig:k_m2d}
}
\caption{(a) and (b): Comparison of varying $\mu$ and constant $\mu$.
We conducted nine experiments for each case.
Note that, the $x$-axis has different meanings for these two cases.
For constant $\mu$, it was fixed to $0.1, ..., 0.9$, respectively, during the entire training.
For varying $\mu$, we linearly changed it from $0$ to $0.1, ..., 0.9$, respectively.
According to results, the varying $\mu$ always achieves higher accuracy than constant $\mu$.
(c) Influence of similarity threshold $\lambda$ on accuracy. When  $\lambda \in [0.5, 0.6]$, the method simultaneously achieves high accuracy on both Market-1501 and DukeMTMC-reID. 
(d) Influence of similarity threshold $\lambda$ on the number of reliable neighborhoods. When $\lambda$ is small, a large number of images are selected. When $\lambda$ is large, only a few images are selected.
(e) and (f): Comparison of adaptive selection (adaptive) and a constant number of neighborhoods (top-$k$) on accuracy. Our adaptive selection outperforms top-$k$ for all cases.
}
\label{Fig:mu_change}
\end{figure*}

\subsection{Ablation Study}

\subsubsection{Exploration for Update Rate $\mu$}
To investigate the effect of $\mu$, different varying rates of $\mu$ were evaluated.
We linearly increased the varying $\mu$ from 0 to 0.1, ..., 0.9, respectively.
Meanwhile, the constant $\mu$ was also used to be compared with the varying $\mu$.
The $\mu$ was fixed to 0.1, ..., 0.9, respectively.
The results are shown in Figure~\ref{Fig:mu_d2m} and Figure~\ref{Fig:mu_m2d}.

First, the varying $\mu$ is superior to the constant $\mu$.
The accuracy curves of the varying $\mu$ are often above those of the constant $\mu$.

Secondly, for varying $\mu$, different varying rates lead to different accuracy.
In Figure~\ref{Fig:mu_d2m} and Figure~\ref{Fig:mu_m2d}, as the varying rates increase, accuracy curves increase at first and then decrease.
In Figure~\ref{Fig:mu_d2m}, when tested on Market-1501, the peak is achieved when $\mu$ is increased from 0 to 0.4.
In Figure~\ref{Fig:mu_m2d}, when tested on DukeMTMC-reID, the peak is achieved when $\mu$ is increased from 0 to 0.5.
\subsubsection{Learning with Adaptive Exploration}

To investigate the effect of learning with adaptive exploration, 
we adopted different values of similarity threshold $\lambda$.
The accuracy is reported in Figure~\ref{Fig:lambda_change}. 
The average number of selected neighborhoods in the last epoch is shown in Figure~\ref{Fig:last_kmean}.
The comparisons between adaptive selection and top-$k$ are shown in Figure~\ref{Fig:k_d2m} and Figure~\ref{Fig:k_m2d}.

In Figure~\ref{Fig:lambda_change}, different values of $\lambda$ result in different accuracy.
With $\lambda$ increasing, accuracy curves (rank-1 accuracy and mAP) increase first and then decrease.
The highest accuracy is achieved when $\lambda$ is equal to 0.55.
The best $\lambda$s for different datasets are similar.
Specifically, high accuracy is achieved when $\lambda = 0.55$ on both Market-1501 and DukeMTMC-reID.

In Figure~\ref{Fig:last_kmean}, when $\lambda = 0.55$, the average number of selected neighborhoods is close to the ground truth.
For example, the Market-1501 training set contains 12,936 images with 751 person identities.
The average number of images per person is 17.2.
When $\lambda = 0.55$, our model selects 18.3 neighborhoods on average in the last epoch, which is close to the ground truth 17.2.

In Figure~\ref{Fig:k_d2m} and Figure~\ref{Fig:k_m2d}, top-$k$ is always under adaptive selection.
For example, when transferring DukeMTMC-reID to Market-1501 in Figure~\ref{Fig:k_d2m}, there exists a large margin between mAP curves of top-$k$ and adaptive selection.
This demonstrates that the adaptive selection can significantly improve the model.

\subsubsection{Learning with Balance}

To investigate the effect of learning with balance, we compared models with balance and models without balance in Table~\ref{tab:target_only}.
Without balance, the model achieves higher accuracy when we set $\delta$ to 0.2 and increased $\mu$ from 0 to 0.4.

Experimental results suggest that the usage of learning with balance significantly improves accuracy.
For example, when transferring DukeMTMC-reID to Market-1501 with CamStyle augmentation, we observe +15.0\% in rank-1 accuracy and +22.9\% in mAP with the balance strategy.

\begin{table}[h]
\footnotesize
\caption{Comparison of our models under different settings on DukeMTMC-reID(Duke) and Market-1501(Market). \textbf{Source-only}: Model is only trained on the source dataset. \textbf{Target-only}: Model is only trained on the target dataset. \textbf{Transfer}: Model is trained on both source dataset and target dataset. \textbf{Cam}: CamStyle \cite{ DBLP:journals/tip/ZhongZZLY19} data augmentation. \textbf{Src.}: Source domain.}
\label{tab:target_only}
\setlength{\tabcolsep}{0.90pt}
\setlength\extrarowheight{1.5pt}
\centering
\vspace{1.0 ex}
\begin{tabular}{c|l|l|c|cccc|c|cccc}
\hline
\multicolumn{3}{c|}{\multirow{2}*{Methods}}   &\multicolumn{5}{c|}{Market-1501 (\%)}    &\multicolumn{5}{c}{DukeMTMC-reID (\%)} \\ \cline{4-13}
\multicolumn{3}{c|}{}                                    &Src.         & rank-1 & rank-5 & rank-10 & mAP               &Src.      & rank-1 & rank-5 & rank-10 & mAP \\ \hline \hline
\multicolumn{3}{c|}{Source-only}              &{Duke}                  &  46.4  &  63.7  &  70.6   & 20.0           &{Market}     &  30.3  &  45.4  &  52.5   & 15.9 \\ \hline

\multirow{4}{*}{\rotatebox[origin=c]{90}{w/o Cam}}   &\multirow{2}{*}{Target-only}       
&adaptive selection                           &\multirow{2}{*}{N/A}     &  35.4  &  54.2  &  63.3   & 14.4           &\multirow{2}{*}{N/A}         &  37.2  &  53.5  &  60.6   & 20.5 \\
&{} 
&adaptive selection + balance                 &{}                       &  57.1  &  75.3  &  81.8   & 34.5           &{}                           &  56.1  &  72.6  &  78.7   & 35.5 \\ \cline{2-13}
&\multirow{2}{*}{Transfer}
&adaptive selection                           &\multirow{2}{*}{Duke}    &  49.6  &  66.0  &  73.0   & 22.4           &\multirow{2}{*}{Market}      &  45.8  &  60.8  &  68.8   & 28.9 \\ 
&{} 
&adaptive selection + balance                 &{}                       &  67.3  &  79.6  &  83.7   & 40.8           &{}                           &  60.1  &  76.3  &  82.3   & 41.8    \\ \hline

\multirow{4}{*}{\rotatebox[origin=c]{90}{w/ Cam}}      &{\multirow{2}{*}{Target-only}}          
&adaptive selection                           &\multirow{2}{*}{N/A}     &  55.4  &  74.8  &  81.6   & 23.4           &\multirow{2}{*}{N/A}         &  42.5  &  57.9  &  64.2   & 19.4  \\
&{} 
&adaptive selection + balance                 &{}                       &  77.5  &  89.8  &  93.4   & 54.0           &{}                           &  63.2  &  75.4  &  79.4   & 39.0 \\ \cline{2-13}
&\multirow{2}{*}{Transfer}
&adaptive selection                           &\multirow{2}{*}{Duke}    &  66.6  &  83.6  &  89.1   & 35.1           &\multirow{2}{*}{Market}      &  59.4  &  72.5  &  78.3   & 36.2 \\
&{}
&adaptive selection + balance                 &{}   &\textbf{81.6}  &\textbf{91.9}  &\textbf{94.6}   &\textbf{58.0}  &{}                           &\textbf{67.9}  &\textbf{79.2}  &\textbf{83.6}   &\textbf{46.7} \\ \hline
                                    
  \end{tabular}
\end{table}

\subsubsection{Comparison between Transfer Learning and Target-only Learning}
In this paper, we considered both domain adaptive re-ID and target-only re-ID of our method.
We reported their results in Table~\ref{tab:target_only}.

Firstly, the source-only model fails to produce good results in the target domain.
For example, when directly transferring DukeMTMC-reID to Market-1501, the source-only model only achieves 46.4\% in rank-1 accuracy and 20\% in mAP.
When directly transferring Market-1501 to DukeMTMC-reID, the source-only model only achieves 30.3\% in rank-1 accuracy and 15.9\% in mAP.
This demonstrates the notorious domain bias problem.

Secondly, our model significantly increases accuracy with transfer learning.
For example, when transferring DukeMTMC-reID to Market-1501, our model achieves 81.6\% in rank-1 accuracy and 58.0\% in mAP. 
Compared with the source-only baseline model, our model increases the accuracy by 35.2\% in rank-1 accuracy and 38.0\% in mAP.
When transferring Market-1501 to DukeMTMC-reID, our model achieves 67.9\% in rank-1 accuracy and 46.7\% in mAP. 
Compared with the source-only baseline model, our model increases the accuracy by 37.6\% in rank-1 accuracy and 20.8\% in mAP.

At last, the target-only learning of our method still achieves competitive accuracy.
For example, on Market-1501, our model with the target-only re-ID protocol achieves 77.5\% in rank-1 accuracy and 54.0\% in mAP. 
Compared with the transfer learning, the target-only learning decreases rank-1 accuracy and mAP by 4.1\% and 4.0\%, respectively.
\subsubsection{Hyper-parameter Analysis}
\begin{figure*}[t]
\centering
\subfigure[Hyper-parameter $\tau$]
{
    \centering
    \includegraphics[width=0.315\textwidth]{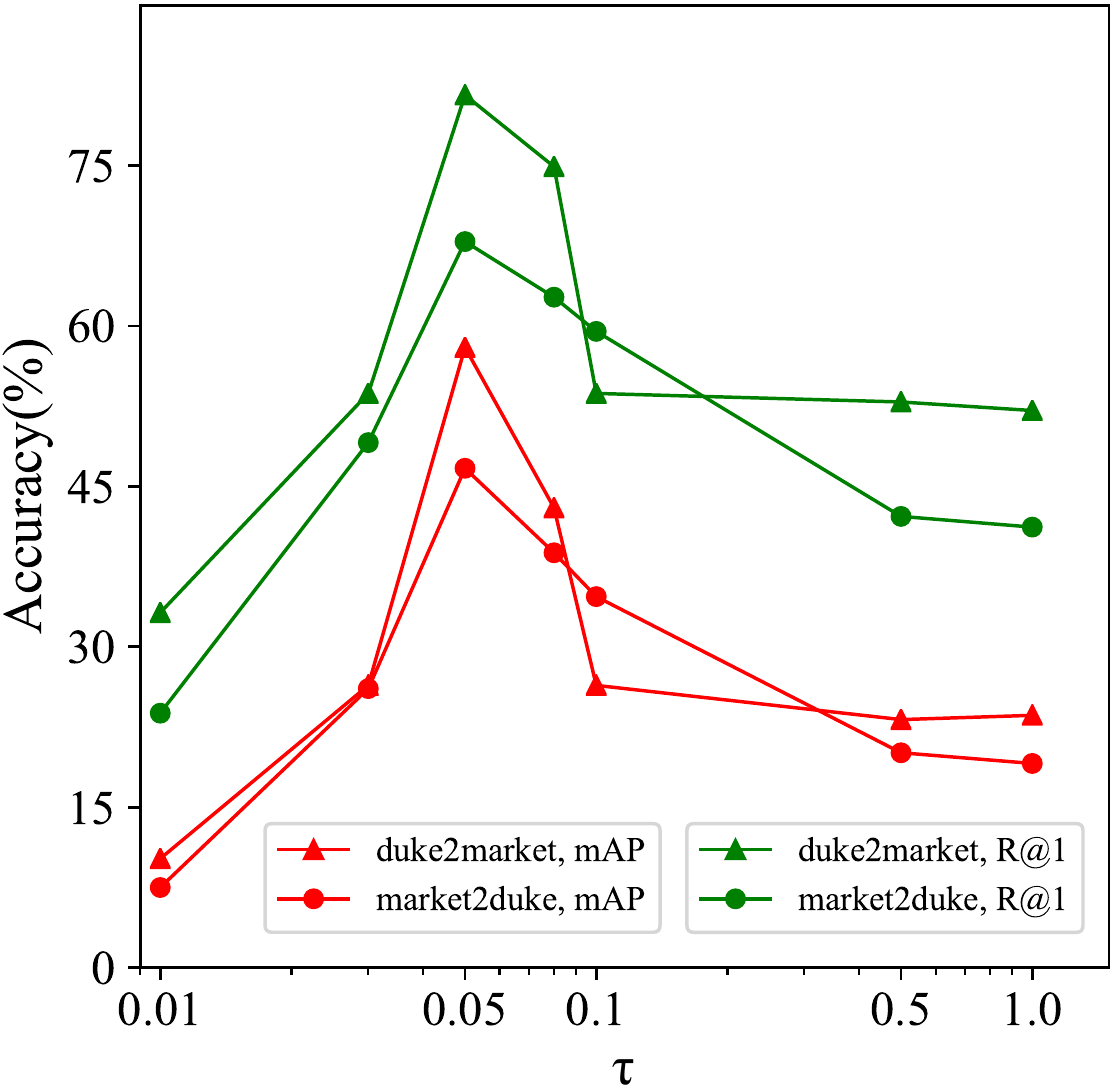}
    \label{Fig:tau}
}
\subfigure[Hyper-parameter $\xi$]
{
    \centering
    \includegraphics[width=0.315\textwidth]{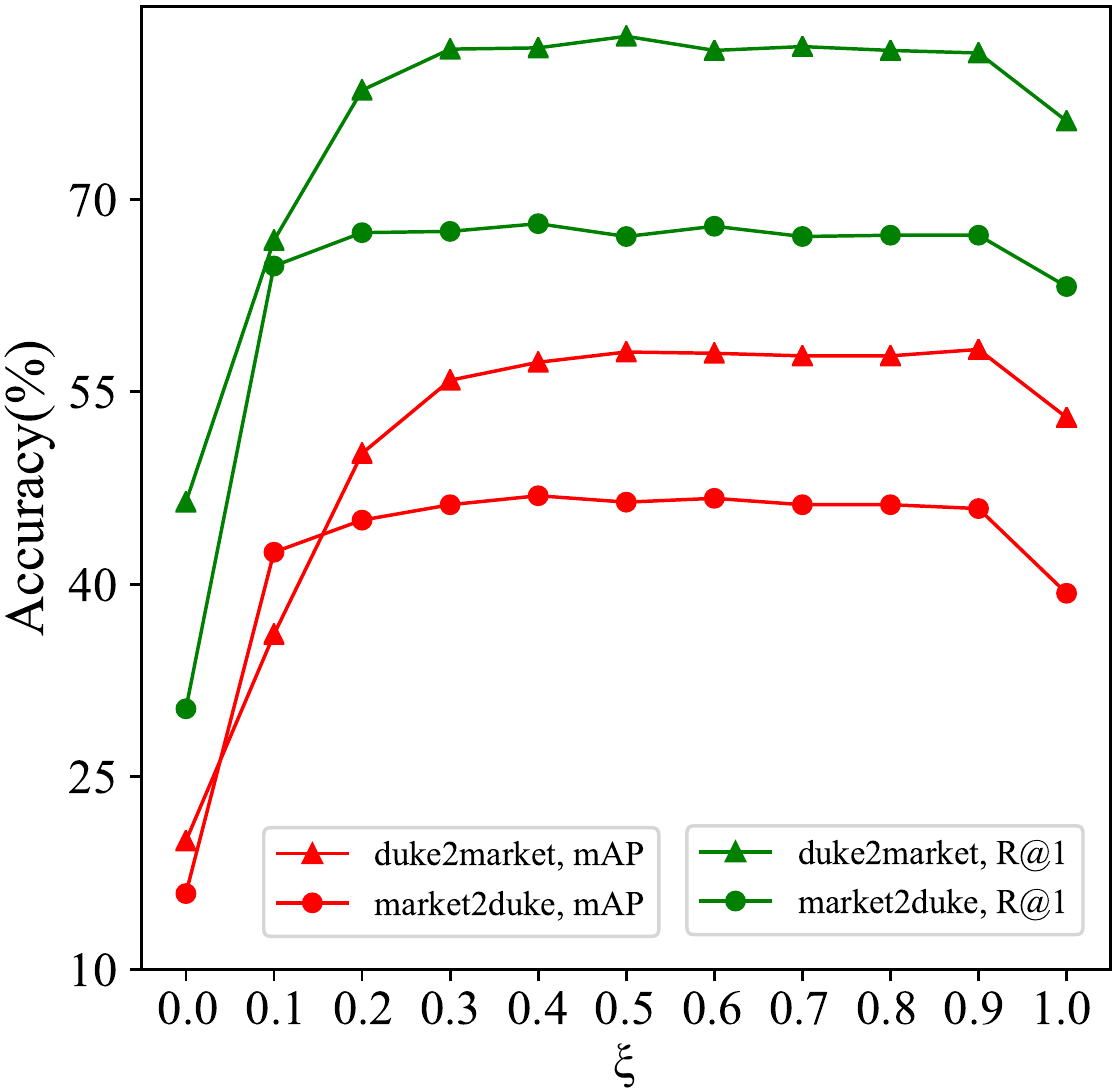}
    \label{Fig:xi}
}
\subfigure[Hyper-parameter $\delta$]
{
    \centering
    \includegraphics[width=0.315\textwidth]{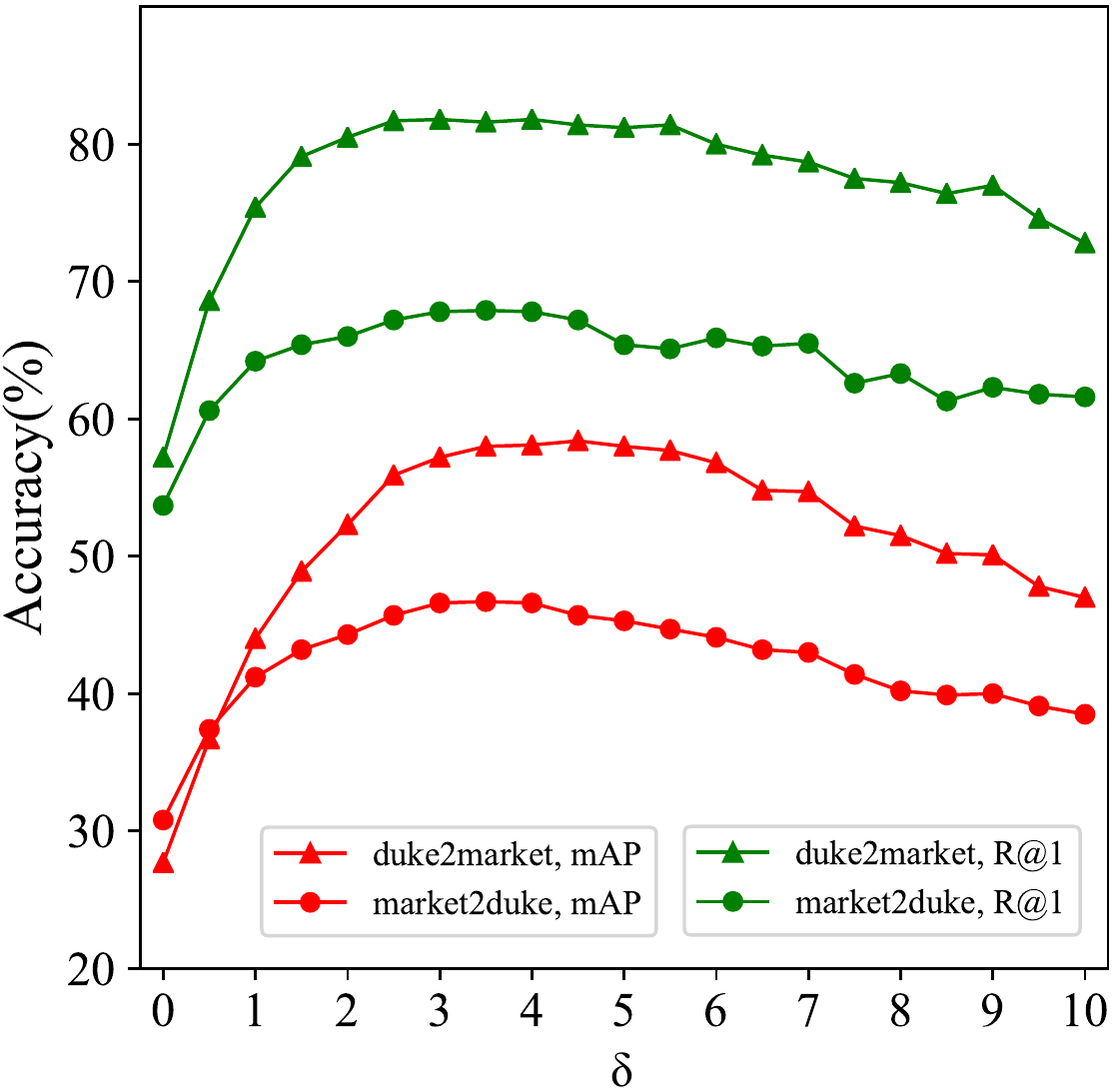}
    \label{Fig:delta}
}
\caption{Impacts of different hyper-parameters.
By default, $\tau$, $\xi$ and $\delta$ are set to 0.05, 0.6 and 3.5, respectively. When evaluating one hyper-parameter, other hyper-parameters are fixed.
}
\label{Fig:parameter analysis}
\end{figure*}
In Figure~\ref{Fig:parameter analysis}, we analyzed the impacts of three hyper-parameters, \textit{i.e.}, $\tau$, $\xi$ and $\delta$.
We varied one hyper-parameter and fixed others to train re-ID models.

In Figure~\ref{Fig:tau}, we changed $\tau$ from 0.01 to 1.0.
As shown in the figure, the higher accuracy is achieved when $\tau$ is equal to 0.05.

In Figure~\ref{Fig:xi}, we changed $\xi$ from 0.0 to 1.0.
When $\xi$ is set to 0.0 or 1.0, the accuracy decreases.
This demonstrates only using source data ($\xi$ = 0.0) or only using target data ($\xi$ = 1.0) achieve lower accuracy than using them together.
Besides, when $\xi \in $ [0.3, 0.9], the accuracy is stable.

In Figure~\ref{Fig:delta}, we changed $\delta$ from 0.0 to 10.0.
When $\delta \in$ [3.0, 4.0], the higher accuracy is achieved.

\subsubsection{Comparison between Triplet and Contrastive Loss}

\begin{table}[t]
\footnotesize
\centering
\setlength{\tabcolsep}{9.7pt}
\setlength\extrarowheight{1.5pt}
\caption{Comparison between triplet loss and contrastive loss}
\label{table: triplet and contrastive loss}

\begin{tabular}{l|cccc|cccc}
\hline
\multirow{2}*{Methods}   &\multicolumn{4}{c|}{DukeMTMC-reID to Market-1501 (\%)}    &\multicolumn{4}{c}{Market-1501 to DukeMTMC-reID (\%)} \\ \cline{2-9}
                                                    & rank-1 & rank-5 & rank-10 & mAP                     & rank-1 & rank-5 & rank-10 & mAP \\ \hline \hline
 Contrastive                    &48.1    &  64.3  &  70.8   & 20.7                    &  35.2  &  51.2  &  57.3   & 19.0 \\ 
 Triplet                  &  56.6  &  74.8  &  81.9   & 28.3                    &  50.2  &  67.1  &  72.8   & 30.7 \\
 Ours                       &  \textbf{81.6}  &  \textbf{91.9}  &  \textbf{94.6}   & \textbf{58.0}                     &  \textbf{67.9}  &  \textbf{79.2}  &  \textbf{83.6}   & \textbf{46.7} \\ \hline
\end{tabular}
\end{table}

In the target domain, the re-ID model is improved by maximizing feature distances between all images and minimizing feature distances between similar images.
Besides the classification framework in AE, triplet loss and contrastive loss can also be used to achieve this objective.
Specifically, for triplet loss, each target image and its neighborhoods are treated as positive pairs while others are viewed as negative samples.
Similarly, for contrastive loss, feature distances between target images and their neighborhoods are minimized while feature distances between target images and the negative samples are maximized.
Compared with AE, as shown in Table~\ref{table: triplet and contrastive loss}, triplet loss and contrastive loss fail to achieve competitive accuracy, demonstrating that our method is more effective than triplet loss and contrastive loss.

\begin{table}[t]
\footnotesize
\centering
\setlength{\tabcolsep}{4.7pt}
\setlength\extrarowheight{1.5pt}
\caption{
Comparison between 2048-dim and 4096-dim features.
\textbf{2048-dim feature}: 2048-dimension features from the Pool-5 layer.
\textbf{4096-dim feature}: 4096-dimension features from the 4096-dimension fully-connected layer.
}
\label{table: 2048 and 4096 dim features}
\begin{tabular}{l|cccc|cccc|c|c}
\hline
\multirow{2}*{Features}   &\multicolumn{4}{c|}{Duke to Market (\%)}    &\multicolumn{4}{c|}{Market to Duke (\%)} & \multirow{2}*{\tabincell{c}{FlOPs\\(E9)}} &\multirow{2}*{\tabincell{c}{Params\\(million)}} \\ \cline{2-9}
                                                    &rank-1 &rank-5 &rank-10 &mAP        &rank-1 &rank-5 &rank-10 &mAP &{} &{}\\ \hline \hline
2048-dim feature                    &81.6    &  91.9  &  94.6   & 58.0                    &  67.9  &  79.2  &  83.6   & 46.7 &2.698 &25.557 \\ 
4096-dim feature                  &  81.3  &  91.5  &  94.1   & 58.5                    &  67.5  &  78.5  &  82.1   & 46.0 &2.715 &33.958 \\ \hline
\end{tabular}
\end{table}

\subsubsection{2048-dim Features vs. 4096-dim Features}
During test, we utilized 2048-dimension features from the Pool-5 layer to retrieve images.
In addition, we also used 4096-dimension features from the 4096-dimension fully-connected layer.
As shown in Table~\ref{table: 2048 and 4096 dim features}, the two features achieve similar accuracy.
However, compared with 2048-dimension features, 4096-dimension features use one more layer for computation, \textit{i.e.}, the 4096-dimension fully-connected layer.
Therefore, as shown in Table~\ref{table: 2048 and 4096 dim features}, more flops and parameters are used for 4096-dimension features.

\subsection{Comparison with State-of-the-art Methods}
\begin{table}[t]
\footnotesize
\centering
\setlength{\tabcolsep}{8.5pt}
\setlength\extrarowheight{1.5pt}
\caption{Comparison with the state-of-the-art methods of \textbf{domain adaptive re-ID} on Market-1501 and DukeMTMC-reID.  $\dagger$ denotes semi-supervised methods. $\ddagger$ denotes direct transfer methods for person re-ID.}
\label{table:state-of-the-art of domain adaptation}

\begin{tabular}{l|cccc|cccc}
\hline
\multirow{2}*{Methods}   &\multicolumn{4}{c|}{DukeMTMC-reID to Market-1501 (\%)}    &\multicolumn{4}{c}{Market-1501 to DukeMTMC-reID (\%)} \\ \cline{2-9}
                                                    & rank-1 & rank-5 & rank-10 & mAP                     & rank-1 & rank-5 & rank-10 & mAP \\ \hline \hline
 PTGAN\cite{DBLP:conf/cvpr/WeiZ0018}                     &  38.6  &   -    &  66.1   &  -                      &  27.4  &   -    &  50.7   &  -  \\ 
 PUL\cite{DBLP:journals/tomccap/FanZYY18}                 &  45.5  &  60.7  &  66.7   & 20.5                    &  30.0  &  43.4  &  48.5   & 16.4 \\ 
 EUG $\dagger$ \cite{DBLP:journals/tip/WuLDYBY19}                  &49.8    &66.4   &72.7  &22.5                         &45.2    &59.2  &63.4    &24.5 \\
 CAMEL\cite{DBLP:conf/iccv/YuWZ17}                       &  54.5  &   -    &   -     & 26.3                    &   -    &   -    &   -     &  -  \\
 DGNet $\ddagger$ \cite{DBLP:journals/corr/abs-1904-07223}          &56.1    &72.2   &78.1  &26.8                         &42.6    &58.6  &64.6    &24.3 \\
 MMFA\cite{DBLP:conf/bmvc/LinLLK18}                       &  56.7  &  75.0  &  81.8   & 27.4                    &  45.3  &  59.8  &  66.3   & 24.7 \\ 
 SPGAN+LMP\cite{DBLP:conf/cvpr/Deng0YK0J18}                 &  57.7  &  75.8  &  82.4   & 26.7                    &  46.4  &  62.3  &  68.0   & 26.2 \\ 
 TJ-AIDL\cite{DBLP:conf/cvpr/WangZGL18}            &  58.2  &  74.8  &  81.1   & 26.5                    &  44.3  &  59.6  &  65.0   & 23.0 \\
 CamStyle\cite{DBLP:journals/tip/ZhongZZLY19}                               &  58.8  &  78.2  &  84.3   & 27.4                    &  48.4  &  62.5  &  68.9   & 25.1 \\
 HHL\cite{DBLP:conf/eccv/ZhongZLY18}               &  62.2  &  78.8  &  84.0   & 31.4                    &  46.9  &  61.0  &  66.7   & 27.2 \\ 
 ARN\cite{DBLP:conf/cvpr/LiYLYDW18}                    &  70.3  &  80.4  &  86.3   & 39.4                    &  60.2  &  73.9  &  79.5   & 33.4 \\ 
 ECN\cite{DBLP:journals/corr/abs-1904-01990}                 &  75.1  &  87.6  &  91.6   & 43.0                    &  63.3  &  75.8  &  80.4   & 40.4 \\ \hline
 Ours                                          &  \textbf{81.6}  &  \textbf{91.9}  &  \textbf{94.6}   & \textbf{58.0}                     &  \textbf{67.9}  &  \textbf{79.2}  &  \textbf{83.6}   & \textbf{46.7} \\ \hline
\end{tabular}
\end{table}
We compared our method with the state-of-the-art unsupervised person re-ID methods on Market-1501, DukeMTMC-reID, and MSMT17.
Experimental results are shown in Table~\ref{table:state-of-the-art of domain adaptation}, Table~\ref{table:state-of-the-art of unsupervised learning}, and Table~\ref{table:MSMT17}, respectively.
Table~\ref{table:state-of-the-art of domain adaptation} and Table~\ref{table:state-of-the-art of unsupervised learning} report the results of the domain adaptive re-ID methods and the target-only re-ID methods, respectively, on Market-1501 and DukeMTMC-reID. 
As for MSMT17, the results are shown in Table~\ref{table:MSMT17}.

In Table~\ref{table:state-of-the-art of domain adaptation}, our method is compared with ten domain adaptive methods of person re-ID.
Compared with the state-of-the-art domain adaptive re-ID method, \textit{i.e.}, ECN \cite{DBLP:journals/corr/abs-1904-01990}, our method increases rank-1 accuracy by 6.5\% and 4.6\% when tested on Market-1501 and DukeMTMC-reID, respectively.

\begin{table}[t]
\footnotesize
\centering
\setlength{\tabcolsep}{8.5pt}
\setlength\extrarowheight{1.5pt}
\caption{Comparison with the state-of-the-art methods of \textbf{target-only re-ID} on Market-1501 and DukeMTMC-reID.}
\label{table:state-of-the-art of unsupervised learning}

\begin{tabular}{l|cccc|cccc}
\hline
\multirow{2}*{Methods}   &\multicolumn{4}{c|}{Market-1501 (\%)}    &\multicolumn{4}{c}{DukeMTMC-reID (\%)} \\ \cline{2-9}
                                                    & rank-1 & rank-5 & rank-10 & mAP                     & rank-1 & rank-5 & rank-10 & mAP \\ \hline \hline
 LOMO\cite{DBLP:conf/cvpr/LiaoHZL15}                     &  27.2  &  41.6  &  49.1   & 8.0                     &  12.3  &  21.3  &  26.6   & 4.8 \\ 
 BOW\cite{DBLP:conf/iccv/ZhengSTWWT15}                   &  35.8  &  52.4  &  60.3   & 14.8                    &  17.1  &  28.8  &  34.9   & 8.3 \\ 
 OIM\cite{DBLP:conf/cvpr/XiaoLWLW17}                       &  38.0  &  58.0  &  66.3   & 14.0                    &  24.5  &  38.8  &  46.0   & 11.3 \\
 BUC\cite{lin2019aBottom}                      &  66.2  &  79.6  &  84.5   & 38.3                    &  47.4  &  62.6  &  68.4   & 27.5 \\ 
 \hline
 Ours (target-only)                        &  \textbf{77.5}  &  \textbf{89.8}  &  \textbf{93.4}   & \textbf{54.0}                     &  \textbf{63.2}  &  \textbf{75.4}  &  \textbf{79.4}   & \textbf{39.0} \\ \hline
\end{tabular}
\end{table}

\begin{table}[h]
\footnotesize
\centering
\setlength{\tabcolsep}{9.4pt}
\setlength\extrarowheight{1.5pt}
\caption{Comparison with the state-of-the-art methods on MSMT17. \textbf{Src.} denotes the source domain. \textbf{*} denotes that the results are reproduced by ourselves.}
\label{table:MSMT17}

\begin{tabular}{l|c|cccc}
\hline
\multirow{2}*{Methods}                           &\multicolumn{5}{c}{MSMT17 (\%)}                                                     \\ \cline{2-6}
                                                 &Src.                          & rank-1 & rank-5 & rank-10 & mAP                    \\ \hline \hline
PTGAN \cite{DBLP:conf/cvpr/WeiZ0018}             &\multirow{3}{*}{Market}       &  10.2  &   -    &  24.4   & 2.9                    \\ 
ECN \cite{DBLP:journals/corr/abs-1904-01990}     &{}                            &  25.3  &  36.3  &  42.1   & 8.5                   \\ 
Ours                                             &{}     &  \textbf{25.5}  &  \textbf{37.3}  &  \textbf{42.6}   & \textbf{9.2}       \\ \hline
PTGAN \cite{DBLP:conf/cvpr/WeiZ0018}             &\multirow{3}{*}{Duke}         &  11.8  &   -    &  27.4   & 3.3                    \\ 
ECN \cite{DBLP:journals/corr/abs-1904-01990}     &{}                            &  30.2  &  41.5  &  46.8   & 10.2                   \\ 
Ours                                             &{}     &  \textbf{32.3}  &  \textbf{44.4}  &  \textbf{50.1}   & \textbf{11.7}      \\ \hline \hline
BUC* \cite{lin2019aBottom}                       &\multirow{2}{*}{N/A}          &  11.5  &  18.6  &  22.3   & 3.4                   \\
Ours (target-only)                               &{}                            &  \textbf{26.6}  &  \textbf{37.0}  &  \textbf{41.7}   & \textbf{8.5}                   \\ \hline

\end{tabular}
\end{table}

In Table~\ref{table:state-of-the-art of unsupervised learning}, we compared our method with four target-only re-ID methods (LOMO\cite{DBLP:conf/cvpr/LiaoHZL15}, BOW\cite{DBLP:conf/iccv/ZhengSTWWT15}, OIM\cite{DBLP:conf/cvpr/XiaoLWLW17} and BUC \cite{lin2019aBottom}).
When tested on Market-1501, LOMO and BOW achieve 27.2\% and 35.8\% in rank-1 accuracy, respectively.
Compared with these methods, the target-only AE method yields the highest accuracy on Market-1501 and DukeMTMC-reID.
Specifically, on Market-1501, AE achieves 77.5\% in rank-1 and 54.0\% in mAP.
On DukeMTMC-reID, AE achieves 63.2\% in rank-1 and 39.0\% in mAP.
Compared with the state-of-the-art target-only re-ID method, \textit{i.e.}, BUC \cite{lin2019aBottom}, AE increases rank-1 accuracy by 11.3\% when tested on Market-1501 and 15.8\% when tested on DukeMTMC-reID.

In Table~\ref{table:MSMT17}, we compared our method with two domain adaptive re-ID methods (PTGAN \cite{DBLP:conf/cvpr/WeiZ0018} and ECN \cite{DBLP:journals/corr/abs-1904-01990}) and one target-only method (BUC \cite{lin2019aBottom}) on MSMT17.

The target-only AE method achieves competitive results on MSMT17 in Table~\ref{table:MSMT17}.
Specifically, the target-only AE method achieves 26.6\% in rank-1 accuracy and 8.5\% in mAP on MSMT17.
Compared with the target-only person re-ID method BUC \cite{lin2019aBottom}, the target-only AE method increases rank-1 accuracy and mAP by 15.1\% and 5.1\%, respectively.

\subsection{Visualization of Feature Space}

\begin{figure}[t]
\centering
\subfigure[without balance]{
    \centering
    \includegraphics[width=0.225\textwidth]{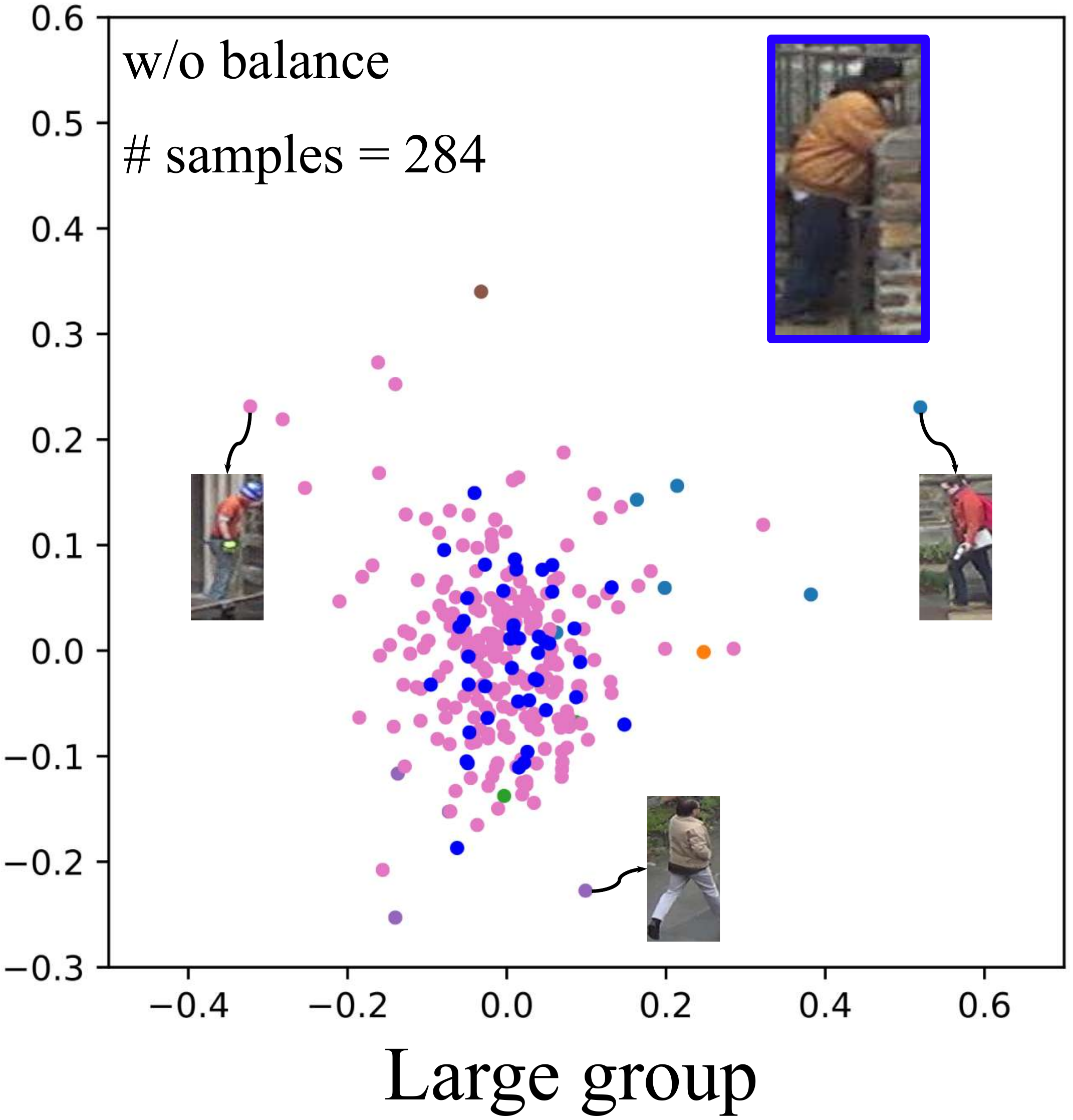}
    \includegraphics[width=0.225\textwidth]{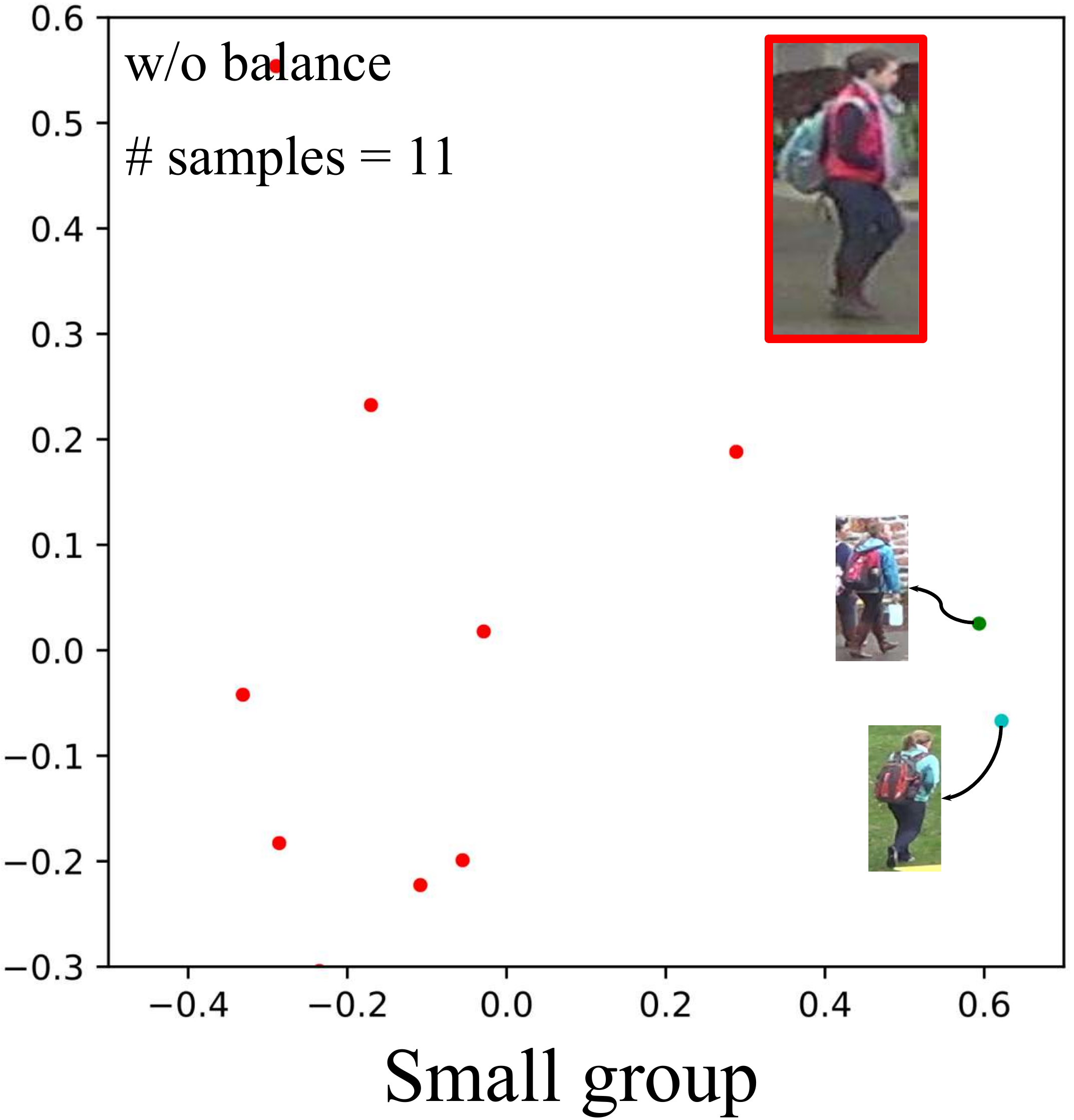}
    \label{Fig:without balance example}
}
\subfigure[with balance]{
    \centering
    \includegraphics[width=0.225\textwidth]{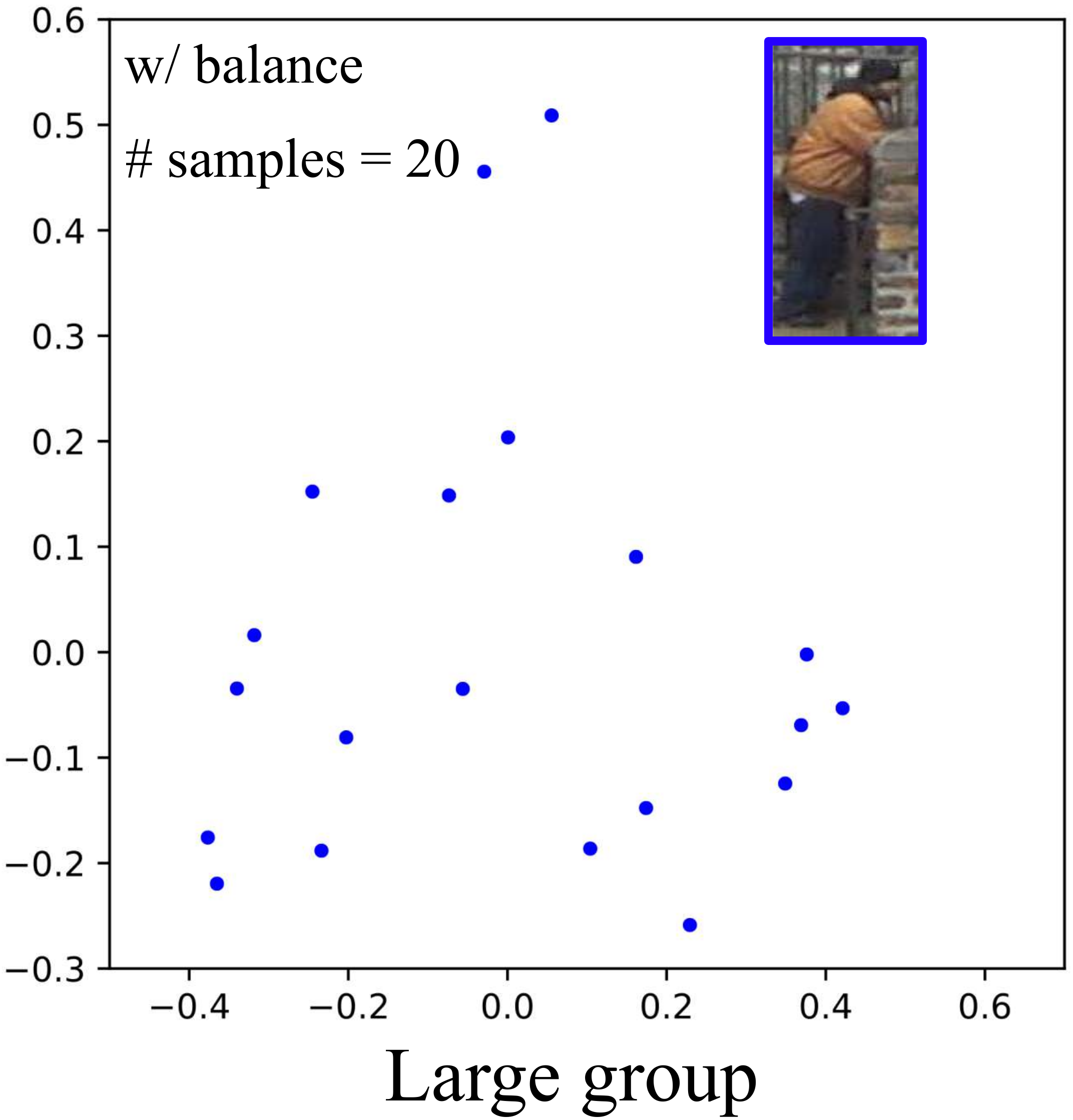}
    \includegraphics[width=0.225\textwidth]{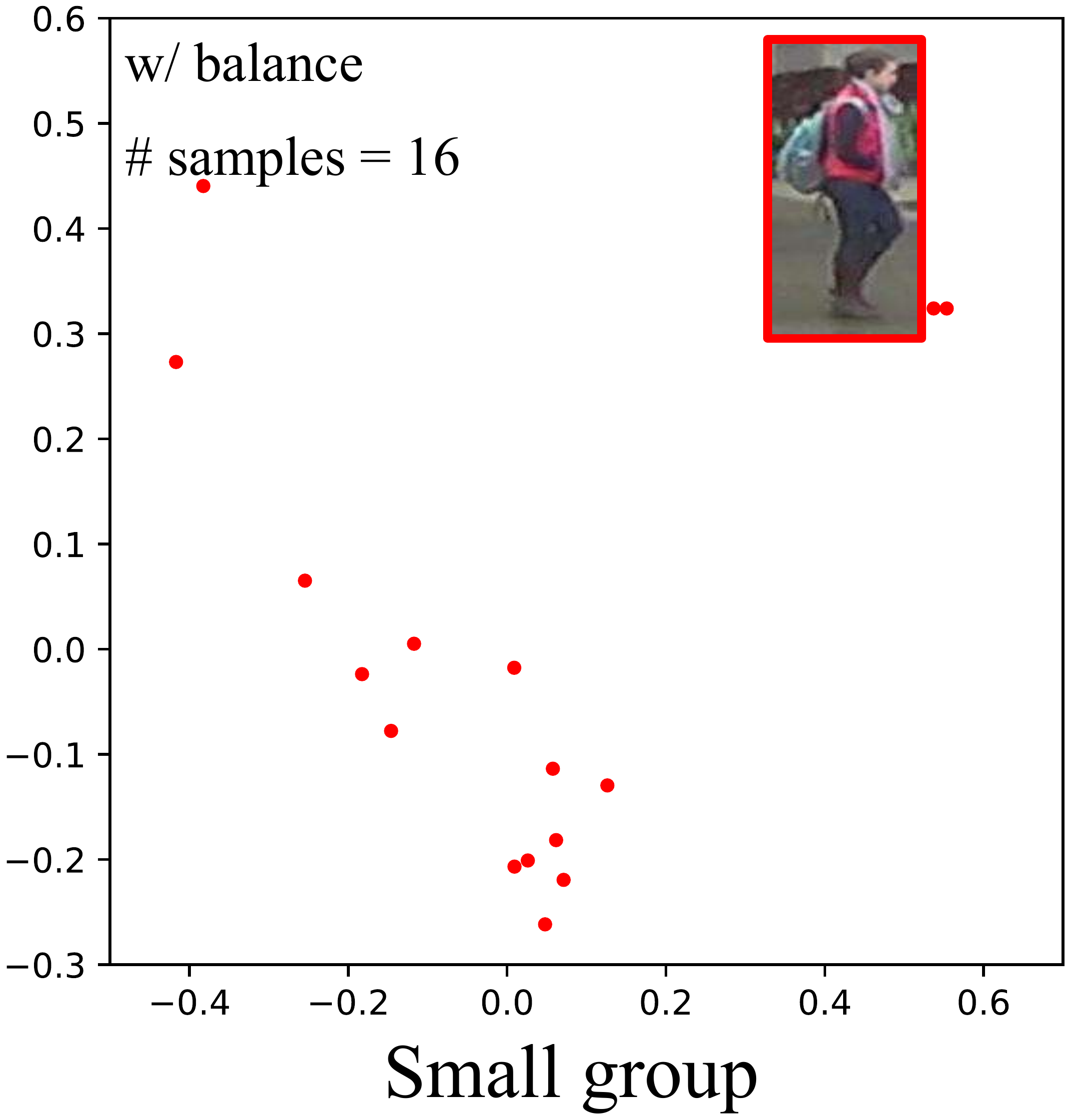}
    \label{Fig:with balance example}
}
\caption{
Visualization for selected neighborhoods according to two images from DukeMTMC-reID. 
The two images (with a blue and red border, respectively) are shown on the upper right.
Each point indicates an image, and different color points indicate different person images.
(a): When learning without balance, one image can have an extremely large number of neighborhoods.
These neighborhoods inevitably contain incorrect persons.
Even for the image in a small group, it still chooses some noisy person images as its neighborhoods.
(b): When learning with balance, the two images select similar number of neighborhoods. 
Meanwhile, these neighborhoods share the same person identities as the two images.
}
\label{Fig:visualization of neighborhoods}
\end{figure}

\subsubsection{Effectiveness of Learning with Balance}
To additionally investigate the effectiveness of learning with balance, we use PCA to visualize neighborhoods selected in the last epoch (60 epoch) by two images on DukeMTMC-reID.
The results are shown in Figure~\ref{Fig:visualization of neighborhoods}.

In Figure~\ref{Fig:without balance example}, without the balance term, one image selects too many neighborhoods while the other one only chooses a few neighborhoods.
Meanwhile, the identities of some neighborhoods are different from those of the two images.
In Figure~\ref{Fig:with balance example}, when learning with balance, the two images enable to select a similar number of neighborhoods.
Also, these neighborhoods share the same identities with the two images.
This indicates that learning with balance manages to help our model classify people accurately.

\begin{figure*}[h]
\centering
\subfigure[Source-only]{
    \centering
    \includegraphics[width=0.23\textwidth]{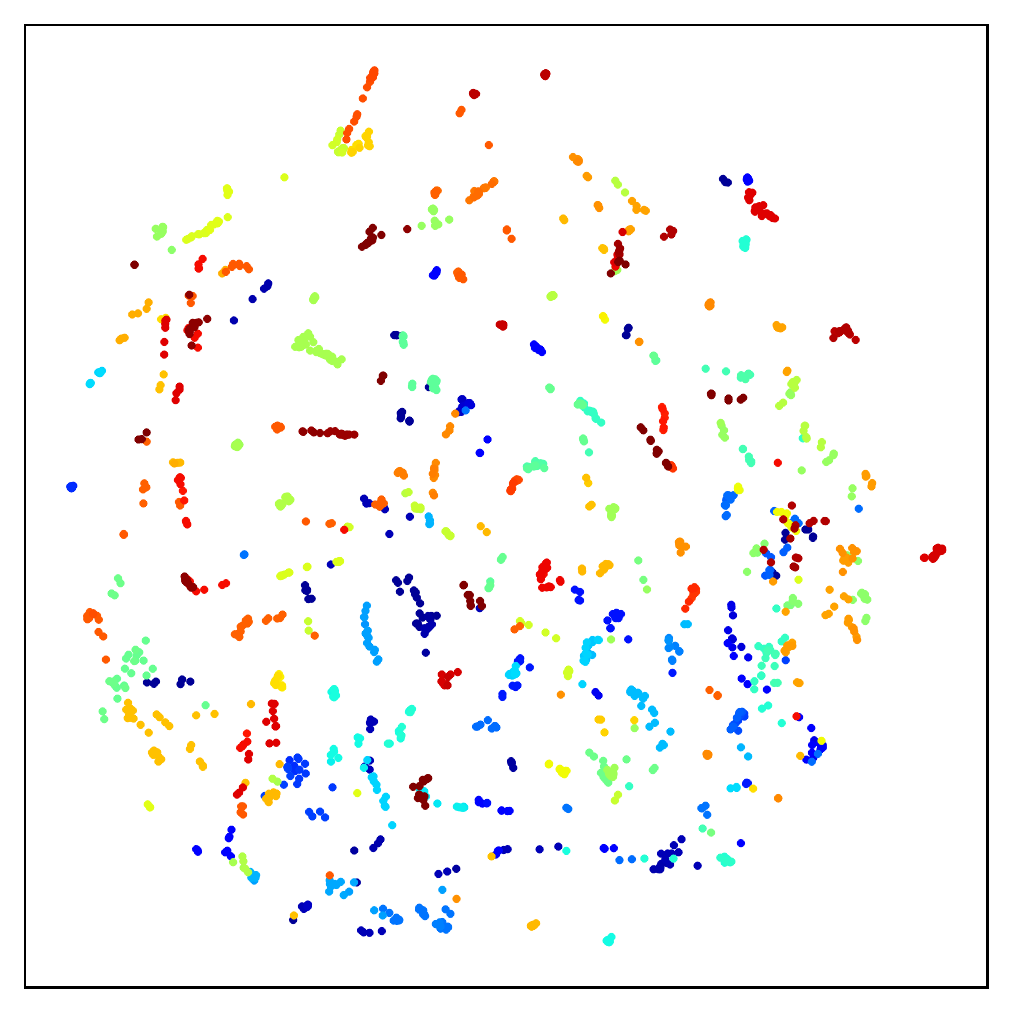}
    \label{Fig:source only tsne}
}
\subfigure[BUC]{
    \centering
    \includegraphics[width=0.23\textwidth]{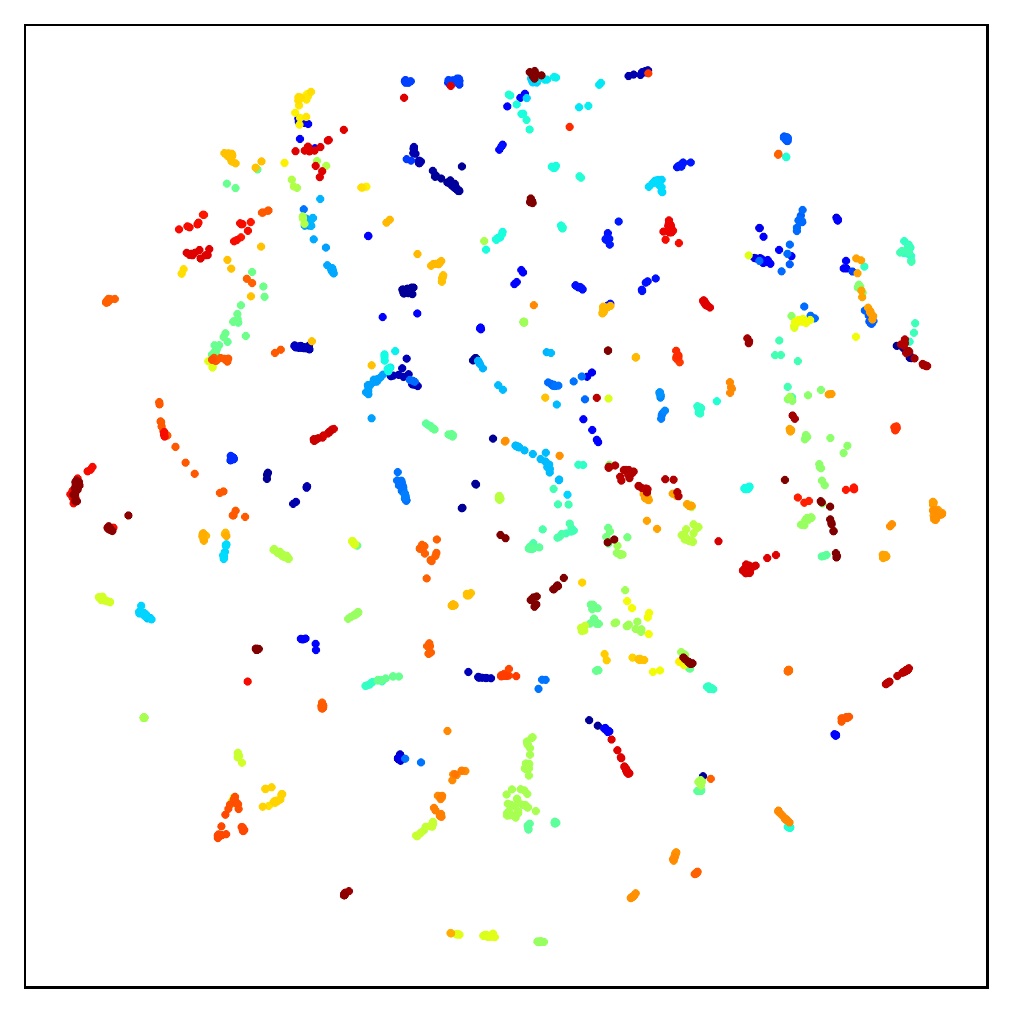}
    \label{Fig:BUC tsne}
}
\subfigure[ECN]{
    \centering
    \includegraphics[width=0.23\textwidth]{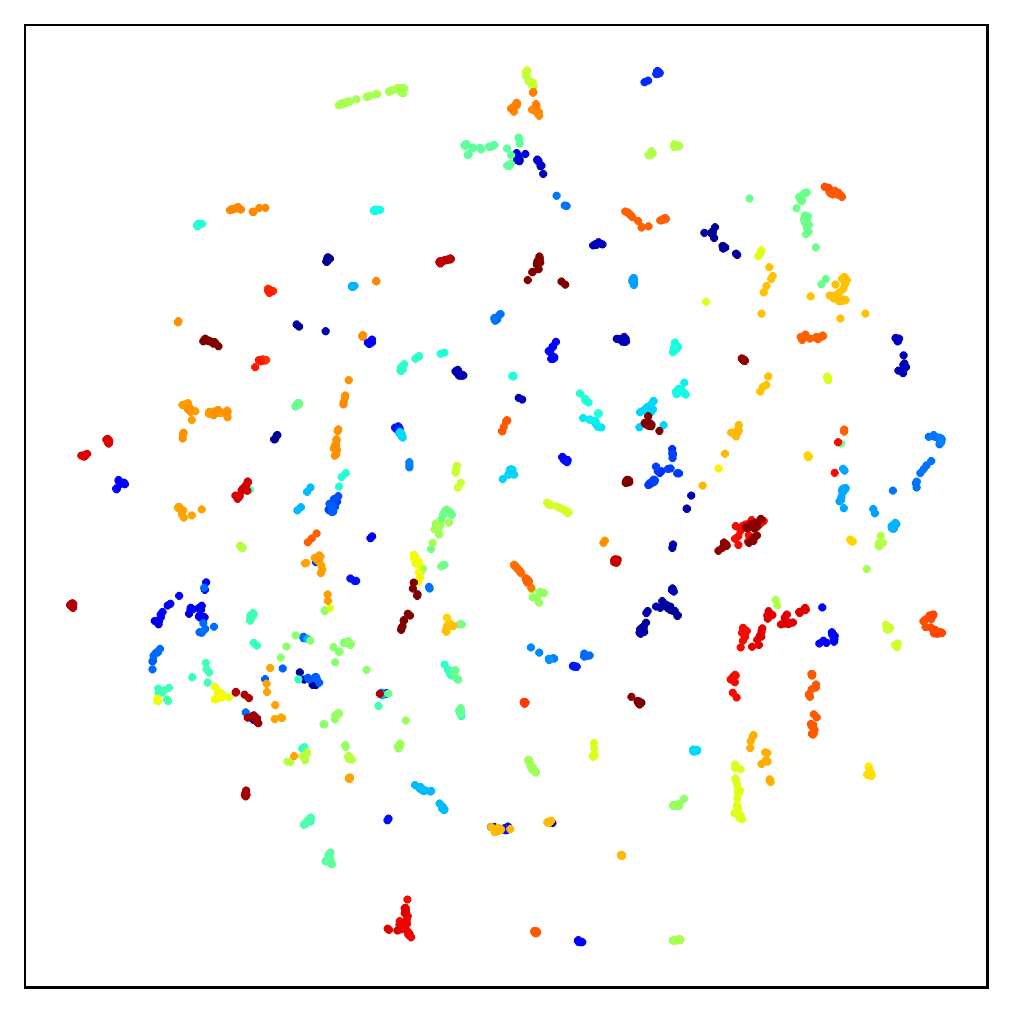}
    \label{Fig:ECN tsne}
}
\subfigure[AE (ours)]{
    \centering
    \includegraphics[width=0.23\textwidth]{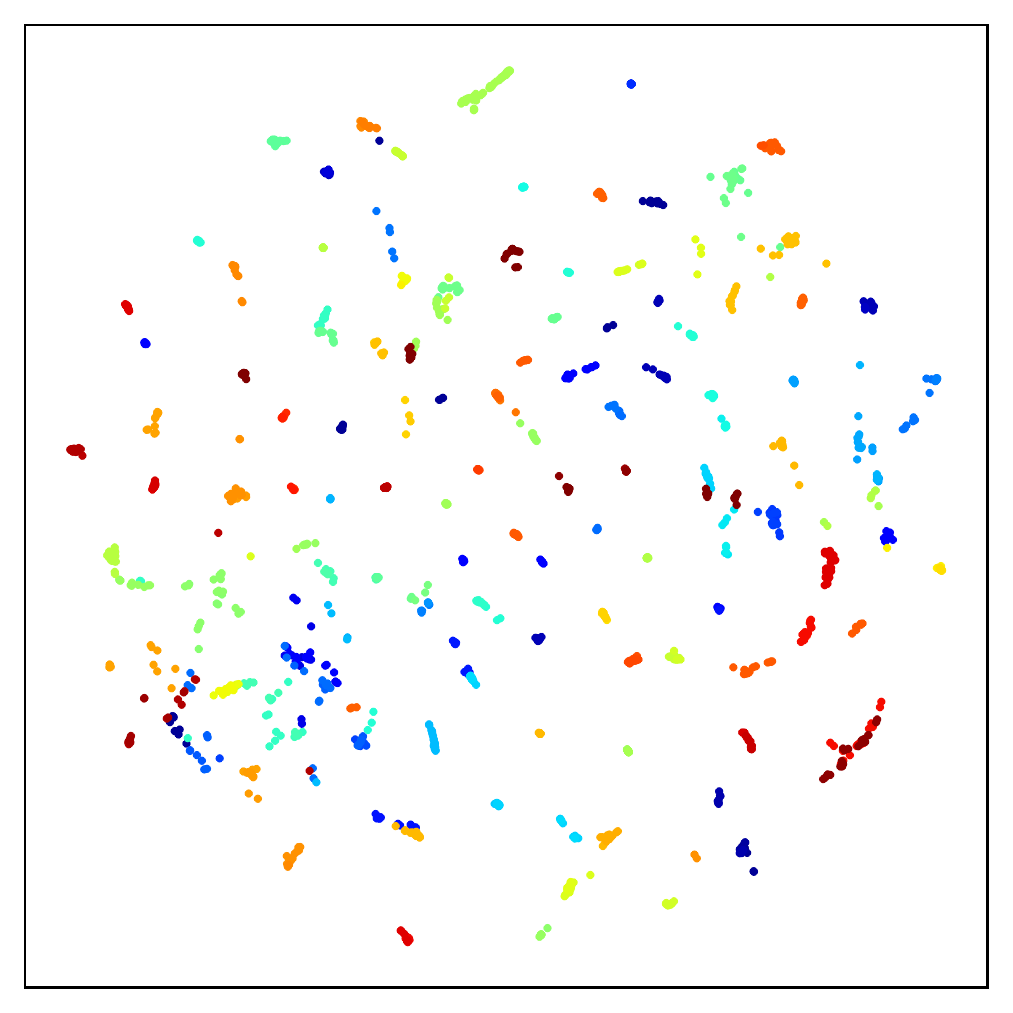}
    \label{Fig:AE tsne}
}
\caption{Visualization for features extracted by source-only, BUC \cite{lin2019aBottom}, ECN \cite{DBLP:journals/corr/abs-1904-01990}, and our AE method.
100 identities with 1,926 images in the gallery of Market-1501 are used. 
Source-only indicates the baseline re-ID model which is only trained on DukeMTMC-reID.
BUC \cite{lin2019aBottom} and ECN \cite{DBLP:journals/corr/abs-1904-01990} are the current best methods of target-only re-ID and domain adaptive re-ID, respectively.
Each point represents an image, and each color represents a person identity.}
\label{Fig:visualization of feature space}
\end{figure*}
\subsubsection{Effectiveness of the AE Method}
To additionally investigate the effectiveness of our method, we use t-SNE \cite{maaten2008visualizing} to visualize feature distributions shown in Figure~\ref{Fig:visualization of feature space}.
Specifically, part of gallery images on Market-1501 (1926 images with 100 identities) are extracted into features and, then the features are projected into a 2-dimension map by t-SNE.
Note that, each point in the map represents one image and points with the same color indicate the same person images.

In Figure~\ref{Fig:AE tsne}, same color points often stay together and are far away from other color points.
This demonstrates our model can extract discriminative features.
Caused by the lack of labels, our model inevitably classifies two similar persons as one identity. 
Therefore, in Figure~\ref{Fig:AE tsne}, there exist two different color points being together. 

We also visualize feature distributions from three other methods, that is, source-only, BUC \cite{lin2019aBottom} and ECN \cite{DBLP:journals/corr/abs-1904-01990} shown in Figure~\ref{Fig:source only tsne}, Figure~\ref{Fig:BUC tsne} and Figure~\ref{Fig:ECN tsne}, respectively.
Source-only indicates the baseline re-ID model fine-tuned on only source data.
BUC \cite{lin2019aBottom} and ECN \cite{DBLP:journals/corr/abs-1904-01990} are the current best methods of target-only re-ID and domain adaptive re-ID, respectively.
Compared with them, our method results in better feature distributions in Figure~\ref{Fig:AE tsne}. 
Specifically, same color points stay closer and fewer different color points stay together by mistake.
This demonstrates the superiority of our method.

\section{Conclusion}
In this paper, we propose the adaptive exploration (AE) method for unsupervised person re-ID.
The AE method explores the unlabeled target domain by considering the feature distances between target images. 
By a non-parametric classifier with a feature memory, AE maximizes distances of all target images and minimizes distances of similar target images.
Meanwhile, we propose to employ a similarity threshold to select reliable similar images.
However, with adaptive selection, some images select too many neighborhoods while others only have a few neighborhoods.
To alleviate the unbalanced problem, we integrate a balance term into the objective loss to prevent images, which have too many neighborhoods, from attracting other images.
As a result, each image tends to select a balanced and reasonable number of neighborhoods.
With the adaptive selection and the balance term, the AE method achieves competitive accuracy on both target-only and domain adaptive re-ID.

\bibliographystyle{ACM-Reference-Format}
\bibliography{acmart}

\end{document}